\documentclass{article}

\usepackage[english]{babel}
\usepackage[shortlabels]{enumitem}
\usepackage{amsmath}
\DeclareMathOperator*{\argmin}{argmin}
\usepackage[letterpaper,top=2cm,bottom=2cm,left=3cm,right=3cm,marginparwidth=1.75cm]{geometry}
\usepackage{float}
\usepackage{algorithm}
\usepackage{algpseudocode}
\usepackage{footnote}
\usepackage{authblk}

\newcommand{\Rho}{\mathrm{P}}

\usepackage{algorithm}
\usepackage{algpseudocode}

\usepackage{etoolbox}
\usepackage{tikz}
\usetikzlibrary{tikzmark}
\usetikzlibrary{calc}

\errorcontextlines\maxdimen

\newcommand{\ALGtikzmarkcolor}{black}
\newcommand{\ALGtikzmarkextraindent}{4pt}
\newcommand{\ALGtikzmarkverticaloffsetstart}{-.5ex}
\newcommand{\ALGtikzmarkverticaloffsetend}{-.5ex}
\makeatletter
\newcounter{ALG@tikzmark@tempcnta}

\newcommand\ALG@tikzmark@start{%
    \global\let\ALG@tikzmark@last\ALG@tikzmark@starttext%
    \expandafter\edef\csname ALG@tikzmark@\theALG@nested\endcsname{\theALG@tikzmark@tempcnta}%
    \tikzmark{ALG@tikzmark@start@\csname ALG@tikzmark@\theALG@nested\endcsname}%
    \addtocounter{ALG@tikzmark@tempcnta}{1}%
}

\def\ALG@tikzmark@starttext{start}
\newcommand\ALG@tikzmark@end{%
    \ifx\ALG@tikzmark@last\ALG@tikzmark@starttext
    \else
        \tikzmark{ALG@tikzmark@end@\csname ALG@tikzmark@\theALG@nested\endcsname}%
        \tikz[overlay,remember picture] \draw[\ALGtikzmarkcolor] let \p{S}=($(pic cs:ALG@tikzmark@start@\csname ALG@tikzmark@\theALG@nested\endcsname)+(\ALGtikzmarkextraindent,\ALGtikzmarkverticaloffsetstart)$), \p{E}=($(pic cs:ALG@tikzmark@end@\csname ALG@tikzmark@\theALG@nested\endcsname)+(\ALGtikzmarkextraindent,\ALGtikzmarkverticaloffsetend)$) in (\x{S},\y{S})--(\x{S},\y{E});%
    \fi
    \gdef\ALG@tikzmark@last{end}%
}

\apptocmd{\ALG@beginblock}{\ALG@tikzmark@start}{}{\errmessage{failed to patch}}
\pretocmd{\ALG@endblock}{\ALG@tikzmark@end}{}{\errmessage{failed to patch}}
\makeatother
\usepackage{indentfirst}

\usepackage{amsmath}
\usepackage{graphicx}
\usepackage[colorlinks=true, allcolors=blue]{hyperref}
\usepackage{subcaption}
\usepackage{amssymb}

\title{Active search for Bifurcations}

\author[1]{Yorgos M. Psarellis\thanks{Currently at Sanofi US}}
\author[1,\textdagger]{Themistoklis P. Sapsis}
\author[2,3,4,$\ddagger$]{Ioannis G. Kevrekidis}
\affil[1]{\small Department of Mechanical Engineering, Massachusetts Institute of Technology, Cambridge, MA, USA}
\affil[2]{Department of Chemical and Biomolecular Engineering, Johns Hopkins University, Baltimore, MD, USA}
\affil[3]{Department of Applied Mathematics and Statistics, Johns Hopkins University, Baltimore, MD, USA}
\affil[4]{Department of Urology, Johns Hopkins University, Baltimore, MD, USA}
\affil[$\dagger$]{Co-corresponding author: \href{mailto:sapsis@mit.edu}{sapsis@mit.edu}}
\affil[$\ddagger$]{Co-corresponding author: \href{mailto:yannisk@jhu.edu}{yannisk@jhu.edu}}

\date{}

\begin{document}
\maketitle

\begin{abstract}
    Bifurcations mark qualitative changes of long-term behavior in dynamical systems and can often signal sudden  (``hard") transitions or catastrophic events (divergences). Accurately locating them is critical not just for deeper understanding of observed dynamic behavior, but also for designing efficient interventions. When the dynamical system at hand is complex, possibly noisy, and expensive to sample, standard  (e.g. continuation based) numerical methods may become impractical. We propose an active learning framework, where Bayesian Optimization is leveraged to discover saddle-node or Hopf bifurcations, from a judiciously chosen small number of vector field observations. Such an approach becomes especially attractive in systems whose state$\times$parameter space exploration is resource-limited. It also naturally provides a framework for uncertainty quantification (aleatoric and epistemic), useful in systems with inherent stochasticity. 
\end{abstract}

\section{Introduction}
\label{sec:into}

What in dynamical systems is called a bifurcation, in a laboratory setting (or in nature) is perceived as a qualitative change in the long-term observed dynamic behavior, sometimes dramatic. Pinpointing the location of these phenomena in state$\times$parameter space, and deciphering the nature of the underlying transitions, has been the focus of significant scientific effort for decades, e.g. in Biology 
\cite{Ferrell2012,ermentrout2010mathematical,Gonze2011, circ_Psarellis2023,Schttler2015,Hastings2018,dushoff1998backwards}) or Chemistry \cite{Uppal1976, Makeev2002, Ess2008, Zhdanov1994, Cassani2021, Lengyel2018, Kevrekidis1986}. In fact, accurate location of bifurcation points remains an active field of research computationally and experimentally \cite{Angeli2004}.
 
When a reliable mathematical model is available, one can locate bifurcations either analytically (if the model is simple enough) or through scientific computing, e.g. in the context of numerical continuation. Such approaches reduce to the numerical solution of a system of (deterministic) equations that characterize bifurcations of a certain type \cite{DOEDEL1991, DOEDEL1991_2, Kuznetsov2004}. Several recently introduced approaches circumvent the need of a closed-form ODE system, and rely solely on sampled data \cite{RICOMARTINEZ20031}. This can be accomplished within the ``Equation-Free" framework \cite{gear2003, Theodoropoulos2000, Makeev2002}, as in Siettos et al. \cite{Siettos2006}. It has been also demonstrated that machine learning (ML) models can be used as surrogates of the hidden ODE system (via nonlinear system identification) and can subsequently be used for numerical continuation or bifurcation detection \cite{Renson2019, Hudson1990, RicoMartinez1994, Lee2023, BEREGI2023109649, McGurk_2023}.

It is also possible to locate bifurcation points in a data-driven way without resorting to numerical continuation of surrogate models. In Anderson et al. \cite{Anderson1999}, a numerical analysis motivated feedback control-based approach was used to simultaneously solve for the bifurcation state vector and the critical parameter value using black-box time-steppers. Such an approach relied on local polynomial fits of the time-stepper iterates to create local surrogate models. Then, feedback control was employed to guide the system directly to bifurcation points. This constitutes an online bifurcation detection (bifurcation continuation) technique that was successfully applied in laboratory settings \cite{RICOMARTINEZ20031}.

In this work, we present an alternative approach, inspired by the field of Machine Learning, specifically active learning. Active learning differs from other ML techniques, in that it allows interaction between the algorithm and the dataset, guiding the sampling of new data in the most informative way for the task at hand \cite{sapsis20, settles2009active, mohamad2018, Pickering2022}. This makes it attractive in applications where data are hard to obtain, such as expensive simulations or difficult laboratory experiments \cite{Blanchard2021b, Greenhill2002, Yang2022}.
In this case, where the objective is the location of a  bifurcation, an active learning protocol is designed 
that iteratively prescribes the next data collection action until convergence to the bifurcation point. Importantly, this approach is probabilistic in nature \cite{Galioto2020}, allowing us to quantify uncertainty and utilize the plethora of search strategies available in probabilistic settings \cite{Blanchard2021, Zhang2023, Blanchard2021_3}.

\begin{figure}[H]
    \centering
    \makebox[0pt][c]{\includegraphics[width=18cm, height=10.125cm] {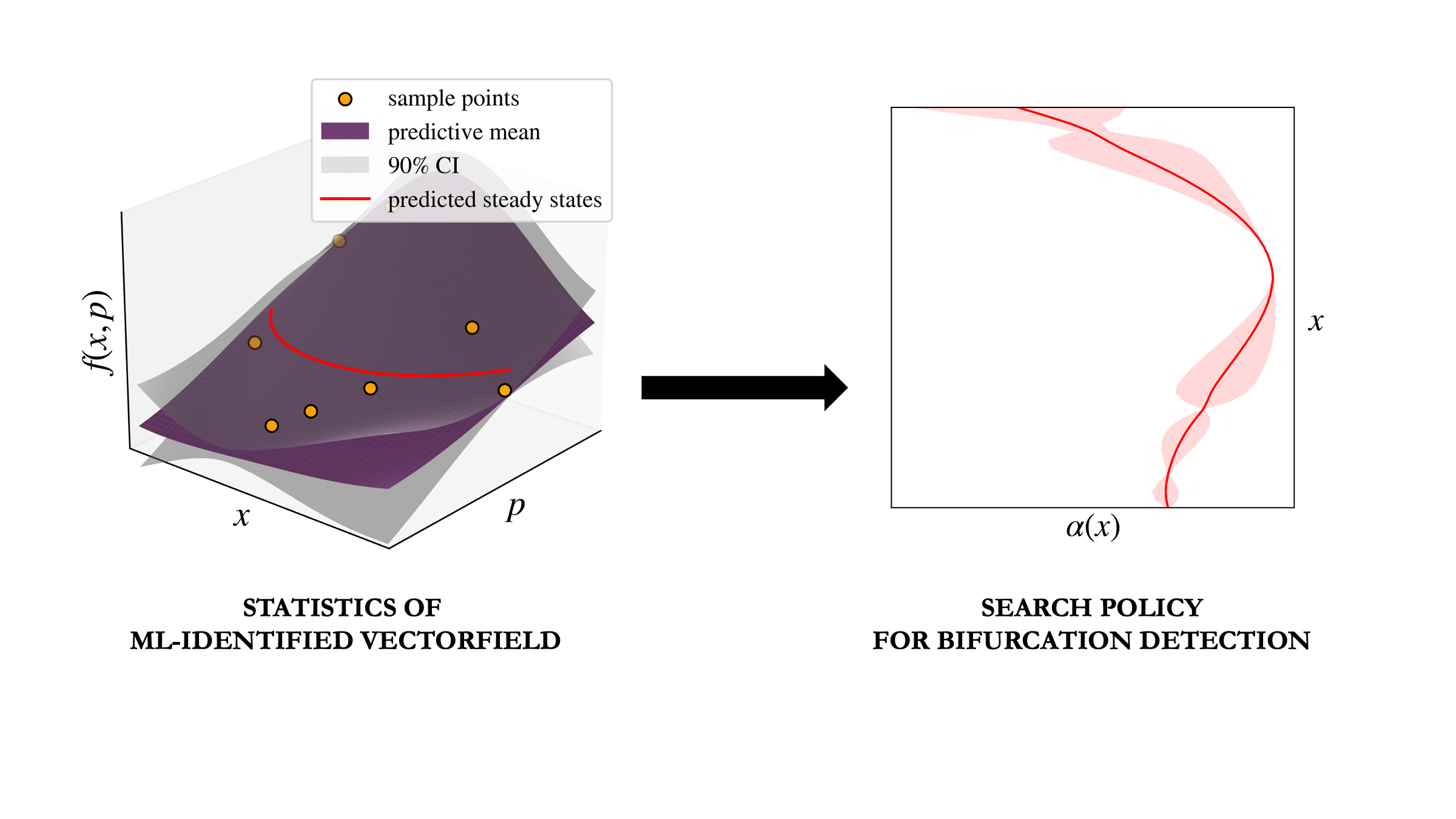}}
    \caption{A schematic summarizing the methodology presented in this manuscript. Here, $x,p$ represent the state variable and parameter respectively, $f(x;p)$ the parameter-dependent vectorfield (see Sec.~\ref{sec:problem}) and $\alpha(x)$ an acquisition function defined on the state space which can be used to guide Bayesian Optimization (see Sec.~\ref{subsec:analytical} for various examples). Solid lines/surfaces represent mean quantities while semi-transparent ones are representative of uncertainty.}
    \label{fig:intro}
\end{figure}

More specifically, we formulate the bifurcation search problem as an optimization problem, where the optimum lies at the bifurcation location. Employing Bayesian Optimization with an adaptively updated Gaussian Process surrogate model, we construct an iterative search protocol embodying an uncertainty-informed strategy. Importantly, we show how uncertainty in ML-predicted vector field values (left panel in Fig.~\ref{fig:intro}) translates to uncertainty in the bifurcation location (right panel in Fig.~\ref{fig:intro}), as described by ML-estimated eigenvalues or derivatives. In the right panel of Fig.~\ref{fig:intro} the chosen acquisition function is plotted in lieu of a bifurcation parameter. In fact, we derive analytical uncertainty expressions that outperform Monte Carlo approximations in computational time, without compromising the accuracy.

The remainder of the manuscript is organized as follows: in Sec.~\ref{sec:problem} bifurcation detection is formulated as an optimization problem, in Sec.~\ref{sec:methods}, all necessary components of the proposed algorithm are presented. In Sec.~\ref{sec:res}, the algorithm's performance is illustrated through various case studies. In Sec.~\ref{sec:conc}
we include the discussion of our results as well as of the work's future outlook.

\section{Problem formulation}
\label{sec:problem}

The objective is to locate bifurcations using (perhaps noisy) vector field observations as the starting point. Consider a set of $N$ initial observations $X=(\mathbf{x}_i,p_i, \mathbf{g}(\mathbf{x}_i;p_i))_{i=1}^N, \hspace{0.2cm} \mathbf{x}\in \mathbb{R}^n \text{ (input variable)}, p\in \mathbb{R}  \text{ (parameter variable)}, \mathbf{g}:\mathbb{R}^{n+1} \rightarrow \mathbb{R}^n$, where $\mathbf{g}(\cdot, \cdot)$ is a (perhaps noisy) observation of the unknown vector field $\mathbf{f}:\mathbb{R}^{n+1} \rightarrow \mathbb{R}^n$ such that $\mathbf{g}(\mathbf{x};p) = \mathbf{f}(\mathbf{x};p)+\epsilon, \hspace{0.2cm} \epsilon\sim \mathcal{N}(0,\sigma_{obs}^2)$.

When the vector field $\mathbf{f}$ is analytically available, one can detect bifurcations by simultaneously solving for a steady state and a criticality condition ($C(\mathbf{x},p)$) which depends on the bifurcation type. The resulting system to be solved is:
\begin{equation} \label{eq:bif_system}
    (\mathbf{x}^b, p^b) = \left\{(\mathbf{x}, p) \hspace{0.2cm} \text{s.t.} \begin{pmatrix}
      \mathbf{f}(\mathbf{x},p) \\
      C(\mathbf{x},p)\\
    \end{pmatrix} =\mathbf{0} \right\}.
\end{equation}

\noindent Alternatively, Eqs.~\ref{eq:bif_system} can be solved as an optimization problem:
\begin{equation} \label{eq:bif_system_opt}
    p^b = \argmin_p\{ C(\mathbf{x}^*,p)^2,  \hspace{0.2cm}\mathbf{x}^* \in  \{\mathbf{x}\in \mathbb{R}^n \hspace{0.2cm}\text{s.t.} \hspace{0.2cm} \mathbf{f}(\mathbf{x},p)=\mathbf{0}\} \}.
\end{equation}

Eqs.~\ref{eq:bif_system} constitute an algebraic system of $n+1$ equations, suitable for finding all $n+1$ unknowns (for a codimension 1 bifurcation), i.e. $n$ components of the state vector as well as the critical parameter value.
In this work, we deal with the following types of bifurcations and corresponding criticality conditions \cite{Kuznetsov2004}:

\begin{itemize}
    \item Fold bifurcation, $n=1$: $C(x,p)=\frac{\partial f}{\partial x}(x,p) $.
    \item Fold bifurcation, $n\geq2$ : $C(\mathbf{x},p)= \min_i|\lambda_i|$, where $\lambda_i$ is the $i-$th eigenvalue of the Jacobian $\mathbf{J}(\mathbf{x};p) \equiv \frac{\partial \mathbf{f}}{\partial \mathbf{x}}$. 
    \item Hopf bifurcation, $n\geq2$ : $C(\mathbf{x},p)= \min_i|Re(\lambda_i)|$, where $\lambda_i$ is the $i-$th eigenvalue of the Jacobian $\mathbf{J}(\mathbf{x};p) \equiv \frac{\partial \mathbf{f}}{\partial \mathbf{x}}.$ Alternative condition: $C(\mathbf{x},p)=tr(\mathbf{J}(\mathbf{x};p))$.
\end{itemize} 

Note that  typically, it is necessary to satisfy several nondegeneracy conditions, involving higher-order derivatives as well \cite{Guckenheimer_1983}. However, we do not consider degenerate cases here. Using only the data available, solving Eqs.~\ref{eq:bif_system} or \ref{eq:bif_system_opt} is not straightforward. Even when we can fit a good model, it should allow the reliable estimation of derivatives, Jacobians and eigenvalues. However, when the data are scarce (e.g. expensive to obtain), identifying a good model is out of reach; equivalently, entire families of models are equally ``good". This necessitates a systematic way to quantify uncertainty not only for $\mathbf{f}(\mathbf{x}, p)$ but also for $C(\mathbf{x}, p)$.

\section{Methodology}
\label{sec:methods}

\subsection{Bayesian Optimization}
\label{subsec:bo}

Bayesian Optimization (BO) is an active learning algorithm suitable for black-box, derivative-free optimization for expensive-to-evaluate functions \cite{KushnerBO, Blanchard2021_3, Blanchard2021, Yang2022, Psarellis2023}. At the BO core lies a surrogate model for the objective function, providing not only predictions, but also uncertainty estimates thereof. The Bayesian nature of the algorithm stems from a user-defined prior distribution that captures our beliefs about the behavior of the unknown objective function. The surrogate model is iteratively improved as more observations are obtained, through Bayesian posterior updating \cite{BO_review}. At every iteration, the next point to be sampled is decided by minimizing an \textit{acquisition function}. Importantly, the choice of the acquisition function reflects a preferred strategy, or a balance between \textit{exploration} and \textit{exploitation} \cite{Brochu2010ATO}. For a generic BO iterative procedure, refer to Alg.~\ref{alg:bo_generic}.

\begin{algorithm}[H]

\caption{Bayesian Optimization}\label{alg:bo_generic}

\textbf{Input}:\\ Observation function of the objective: $g(\mathbf{x})=f(\mathbf{x}) + \epsilon, \hspace{0.2cm} \epsilon \sim \mathcal{N}(0, \sigma_{obs}^2)$, \\Initial dataset: $X_0=(\mathbf{x}_i, g(\mathbf{x}_i))_{i=1}^N$, \\ 
Choice (and parametrization) of the acquisition function: $a(\cdot)$\\
Evaluation budget: $N_{budget}$ \\

\textbf{Output}: \\Estimated optimum: $\mathbf{x}^*$
\vspace{0.2cm}

\begin{algorithmic}

\State $i\gets 0$

\While{ $i \leq N_{budget}$}:
\State Fit surrogate model $\hat{g}\approx f (\mathbf{x})$,  given $X_i$.
\State  Minimize the acquisition function: $\mathbf{x}^{new} \leftarrow \argmin_{\mathbf{x}} a(\mathbf{x}; \hat{g} ,X_i)$
\State  Observe the objective function at the new point: $g^{new} \leftarrow g(\mathbf{x}^{new})$
\State  Augment the dataset with the new sample: $X_{i+1} \leftarrow X_i \cup (\mathbf{x}^{new}, g^{new})$
\State $i \gets i+1$
\EndWhile

\end{algorithmic}
\end{algorithm}

Gaussian Processes have been especially popular as a choice for surrogate models \cite{RasmussenGP}, as they naturally allow for uncertainty quantification of the predictions. However, ensemble neural networks or dropout neural networks can also be employed, especially for higher-dimensional inputs \cite{Pickering2022, Guo2021, guth2023evaluation}.


\subsection{Gaussian Process Regression}
\label{subsec:gp}

In this work, the surrogate models of choice are Gaussian Processes (GPs).
They can be understood as distributions over the space of functions with a Gaussian prior, and can be fully specified by a mean function $m(\cdot)$ (usually set to zero during preprocessing) and a covariance function $k(\cdot, \cdot)$ \cite{RasmussenGP, Lee2023_2}. The covariance function is commonly formulated by a Euclidean-distance kernel function in the input space \cite{psarellis2022datadriven}. Given noisy observations $y = f(\mathbf{x})+\epsilon, \epsilon \sim \mathcal{N}(0, \sigma_{obs}^2)$ and an input-output dataset $D = (\mathbf{X}, \mathbf{y})$ written in matrix form ($\mathbf{y}$ consists of elements $y = f(\mathbf{x})+\epsilon$):
\begin{equation} \label{gp_predic1}
    \left[ \begin{array}{l}
    \mathbf{y} \\
    \mathbf{y^*}  \end{array} \right]  \sim \mathcal{N} \left(\mathbf{0}, 
    \left[ \begin{array}{cc}   \mathbf{K}+ \sigma_{obs}^2 \mathbf{I} & \mathbf{K}_*  \\   \mathbf{K}^\top_* & \mathbf{K}_{**}  \end{array}  \right]\right),
\end{equation} 

\noindent where $\mathbf{y}^*$ is a predictive distribution for test data $\mathbf{X}^*$, $\mathbf{K}$ represents a covariance matrix between training data, $\mathbf{K}_*$ represents a covariance matrix between training and test data, while $\mathbf{K}_{**}$ represents a covariance matrix between test data.
Conditioning on the training data, we can formulate the predictive distribution:
\begin{align} \label{gp_predic2}
    &  \mathbf{y^*} \sim \mathcal{N}( \mathbf{K}_*(\mathbf{K}+\sigma_{obs}^2 \mathbf{I})^{-1} \mathbf{y},  \mathbf{K}_{**}- \mathbf{K}^\top_*(\mathbf{K}+\sigma_{obs}^2 \mathbf{I})^{-1} \mathbf{K}_*).
\end{align} 
Here, we use the standard squared exponential kernel with automatic relevance determination \cite{RasmussenGP}. The hyperparameters of $k(\cdot, \cdot)$ are found through maximum likelihood estimation, resulting in posterior distributions \cite{psarellis2022datadriven}. Importantly, the tractability and normality of the posterior distribution (Eq.~\ref{gp_predic2}) allows the derivation of closed-form derivatives/jacobians of the predictions and their uncertainty \cite{NEURIPS2018_c2f32522}.

\subsection{Active learning for bifurcation detection}
\label{subsec:formulation}

Instead of solving the augmented set of  Eqs.~\ref{eq:bif_system} or \ref{eq:bif_system_opt}, we utilize the Bayesian Optimization framework, described in Sec.~\ref{subsec:bo} to locate the bifurcations. Specifically, we first locate a 1D solution branch (i.e. solutions of $\mathbf{f}(\mathbf{x},p) = 0$ as in Eq.~\ref{eq:bif_system_opt}), expressed w.r.t. an appropriate independent variable. Then, we seek to minimize $C(\mathbf{x},p)^2$ over the 1D steady state branch. Using ideas from uncertainty quantification, we show how uncertainty estimates, provided by the surrogate model, propagate to uncertainty of the steady states,  steady state derivatives, the respective Jacobian matrices and their eigenvalues. A summary of such expressions, and the resulting acquisition functions can be found in Sec.~\ref{subsec:analytical} and the detailed derivation thereof in the Appendix (Sec.~\ref{sec_app:analytical}).

To illustrate this approach, we refer to Alg.~\ref{alg:nd_hopf} that deals with the detection of Hopf bifurcations.
This algorithm can be easily adapted for the search of other bifurcations described in this section, with special care to change the objective function accordingly. Some other important practical considerations include:

\begin{enumerate}[(i)]
    \item It is crucial to choose a single appropriate independent variable to parametrize the  one-dimensional solution branch. In the case of Hopf bifurcations, the bifurcation parameter is the natural choice. In the case of fold bifurcations, however, the implicit function theorem prohibits us from expressing the steady states as a function of the bifurcation parameter \cite{krantz2002implicit}. In fact, it is more convenient to chose one of the state variables as the independent variable. Note, that the same issue appears in numerical continuation, and the independent variable is usually chosen to be the arclength (i.e. pseudo-arclength continuation) \cite{DOEDEL1991, DOEDEL1991_2}. 
    \item It is not always possible to numerically solve the surrogate model for steady states, especially when the model is not reliable, e.g. right upon initialization. In these instances, we resort to uncertainty sampling  \cite{Gramacy2009} until the surrogate is updated to the point where steady states can be reliably found. See Sec.~\ref{subsec_app:us} in the Appendix for a sample implementation of uncertainty sampling.
    \item The analytical formulas (summarized in Sec.~\ref{subsec:analytical} and in detail shown in the Appendix, Sec.~\ref{sec_app:analytical}) quantifying the uncertainty of steady states, derivatives/Jacobians and eigenvalues are approximations based on the assumption of small uncertainty of the approximated vector field. As a consequence, such approximations are accurate only asymptotically. In practice, however, they turn out to perform very well, even in the initial iterations, when uncertainty of the vector field is important.
\end{enumerate}

\begin{algorithm}

\caption{BO search for Hopf bifurcations in 2D}\label{alg:nd_hopf}

\textbf{Input}:\\ Obervation function of the vector field $\mathbf{g}(\mathbf{x}; p)= \mathbf{f}(\mathbf{x}; p) + \epsilon, \hspace{0.2cm} \epsilon \sim \mathcal{N}(0, \sigma_{obs}^2)$, \\Initial dataset $X_0=(\mathbf{x}_i,p_i, \mathbf{g}(\mathbf{x}_i;p_i))_{i=1}^N$, \\ 
BO termination criteria: evaluation budget $N_{budget}$, absolute convergence tolerance $\epsilon_{conv}$, \\
Objective function: $F(x,p) = \left(\min_i|Re(\lambda_i)|\right)^2$\\

\textbf{Output}: \\Estimated bifurcation parameter $p^b$ and state $\mathbf{x}^b$.
\vspace{0.2cm}

\begin{algorithmic}

\Procedure{Bayesian Optimization}{}

\State $\delta \gets 100$
\State $(\mathbf{x}^{old}, p^{old}) \gets (\mathbf{x}_{N} , p_{N})$

\While{$i<N_{budget}$ and $\delta>\epsilon_{conv}$}
\State Train a Gaussian Process $\mathbf{\hat{g}} \approx \mathbf{f}(\mathbf{x}, p)$, given dataset $X_i$,
\State with $\mathbf{\hat{g}}_\mu(\cdot, \cdot), \mathbf{\Sigma}(\cdot, \cdot)$ the pointwise predictive mean and variance.
\State  Minimize the acquisition function.
\State \hspace{1cm} $p^{new} \gets argmin_{p}(\alpha(p); \mathbf{\hat{g}})$
\State Given the steady state distribution $\mathcal{N}(\mathbf{x}^*_\mu, \mathbf{\Sigma}^*)$ at $p^{new}$, sample a single realization. 
\State \hspace{1cm} $\mathbf{x}^{*,new} \gets \mathcal{N}(\mathbf{x}^*_\mu, \mathbf{\Sigma}^*)$
\State Observe the vector field at $(\mathbf{x}^{*,new}, p^{new})$.
\State \hspace{1cm}  $g^{new} \gets \mathbf{f}(\mathbf{x}^{*,new}, p^{new}) + \epsilon$
\State Augment dataset $X$ with the new sample.
\State \hspace{1cm} $X_{i+1} \gets X_i \cup (\mathbf{x}^{*,new}, p^{new}, g^{new})$
\State Check covergence.
\State\hspace{1cm}  $\delta \gets  \sqrt{(p^{new}-p^{old})^2+ ||\mathbf{x}^{new} -\mathbf{x}^{old}||_2^2}  $
\State $(\mathbf{x}^{old}, p^{old}) \gets (\mathbf{x}^{*,new} , p^{new})$
\State $i \gets i+1$

\EndWhile
 \State $(\mathbf{x}^{b}, p^{b}) \gets (\mathbf{x}^{*,new} , p^{new})$
\EndProcedure

\vspace{0.4cm}

\Procedure{Acquisition function: $\alpha(p; \mathbf{\hat{g}})$}{}
\State Given an initial state $\mathbf{x}_0$, numerically solve $\mathbf{\hat{g}}_{\mu}$ for the steady state of the predictive mean.  
\State  \hspace{1cm} $\mathbf{x}^*_\mu(p) \xleftarrow{Newton} \mathbf{x}_0,p\hspace{0.2cm}$

\State Calculate the variance of such steady states, given $\mathbf{J_\mu}$, the Jacobian of $\mathbf{\hat{g}_\mu}$.

\State \hspace{1cm} $\mathbf{\Sigma^*} \leftarrow (\mathbf{J}_\mu^{-1}(\mathbf{x}_\mu^*,p)) \mathbf{\Sigma}(\mathbf{x}_\mu^*,p)  (\mathbf{J}_\mu^{-1}(\mathbf{x}_\mu^*,p))^\top $

\State  Calculate the covariance of the Jacobian  $\mathbf{\Sigma_{J}}$ at the steady state (see Appendix, Sec.~\ref{subsubsec_app:eighopf}).

\State  Calculate the predictive mean eigenvalue with the minimum absolute real part.

\State \hspace{1cm} $E[Re\{ \lambda(p)\}] \gets \lambda_k(p), k = argmin_k|Re\{\lambda_k(\mathbf{J_\mu}(\mathbf{x}_\mu^*,p))\}|$

\State Calculate the variance of the real part of that eigenvalue $Var(Re\{ \lambda(p)\})$, 

\State given the variance of the Jacobian and the left and right eigenvectors of $\mathbf{J}_\mu$
\State (see Appendix, Sec.~\ref{subsubsec_app:eigv})


\State Calculate the statistics of the squared real part of the eigenvalue of interest

\State \hspace{1cm} $E[Re\{ \lambda(p)\}^2] \gets E[Re\{ \lambda(p)\}]^2 + Var(Re\{ \lambda(p)\}) $
\State \hspace{1cm} $Var(Re\{ \lambda(p)\}^2) \gets 2Var(Re\{ \lambda(p)\})(Var(Re\{ \lambda(p)\})+2E[Re\{ \lambda(p)\}]^2) $
\State  Calculate a standard acquisition function $\alpha(p; \mathbf{\hat{g}})$, using the above statistics,
\State e.g. Lower Confidence bound:
\State \hspace{1cm} $\alpha(p; \mathbf{\hat{g}}) = E[Re\{\lambda(p)\}^2]-\beta \sqrt{Var(Re\{ \lambda(p)\}^2)}$
\EndProcedure

\end{algorithmic}
\end{algorithm}

The proposed Alg.~\ref{alg:nd_hopf} can be validated via Monte Carlo sampling. In this case, all statistics (mean and variance of steady states, mean and variance of Jacobians, mean and variance of the eigenvalues) are estimated by a sample of realizations of the GP surrogate (See Sec.~\ref{subsec_app:mc} in the Appendix). In Sec.~\ref{sec:res} we juxtapose the performance of Alg.~\ref{alg:nd_hopf} with Monte Carlo sampling of various sample sizes to highlight accuracy and computational time benefits.

\subsection{The case of a black-box time stepper}
\label{subsec:timestepper}
In the case where vector field measurements are available (even noisy), the approach presented in Sec.~\ref{subsec:formulation} is the natural choice. It is also applicable when we can only observe the state after a short time interval $\Delta t$ upon initialization, for example assuming:
\begin{equation} \label{eq:euler_stepper}
    \mathbf{x}(t+\Delta t) \approx \mathbf{x}(t) + \Delta t \mathbf{f}(\mathbf{x}(t);p) \Rightarrow \mathbf{f}(\mathbf{x}(t);p) \approx  \frac{\mathbf{x}(t+\Delta t) - \mathbf{x}(t)}{\Delta t},
\end{equation}
\noindent where $\Delta t$ is the (known) timestep.
In the cases, however, where Eq.~\ref{eq:euler_stepper} is not a good approximation, subsequent state vectors appear as iterates of a black-box timestepper:
\begin{equation}
     \mathbf{x}(t+\Delta t) = \mathbf{h}(\mathbf{x}(t);p),
\end{equation}
\noindent where $\mathbf{h}:\mathbb{R}^{n+1}\rightarrow\mathbb{R}^n$ is some unknown map.
In this case, bifurcation detection is enabled through the following modifications in the criticality conditions:
\begin{itemize}
    \item Fold bifurcation, $n=1$: $C(x,p)=\frac{\partial h}{\partial x}(x,p)-1 $
    \item Fold bifurcation, $n\geq2$ : $C(\mathbf{x},p)= \min_i|\lambda_i-1|$, where $\lambda_i$ is the $i-$th eigenvalue of the Jacobian $\mathbf{J}(\mathbf{x};p) \equiv \frac{\partial \mathbf{h}}{\partial \mathbf{x}}$ 
    \item Neimark-Sacker bifurcation, $n\geq2$ : $C(\mathbf{x},p)= \min_i|\lambda_i\bar{\lambda_i}-1|$, where $\lambda_i$ is the $i-$th eigenvalue of the Jacobian $\mathbf{J}(\mathbf{x};p) \equiv \frac{\partial \mathbf{h}}{\partial \mathbf{x}}$ and overbar denotes complex conjugate. 
\end{itemize} 
These expressions are equivalent to the ones used in \cite{Anderson1999}, where, however the determinant of the Jacobian was used.

\subsection{The case of  partial differential equations and agent based models}
\label{subsec:abm}
Partial differential equations (PDEs) and agent based models (ABMs) can also be treated within the black-box framework discussed in Secs.~\ref{subsec:formulation},  \ref{subsec:timestepper}. In fact, both PDEs and ABMs can be discretized into a (usually large) number of coupled ODEs. Such an approach involves high-dimensional input and output spaces, which are usually inefficient to work with. A possible solution is to employ reduced order descriptions of PDEs and ABMs,
for example, via expanding the state vector in an appropriately chosen lower-dimensional orthogonal basis \cite{GEAR2002941, Fabiani2024}.

Such as basis can often be further reduced  under the assumption that the leading components ``slave" a number of following components of the chosen basis near the bifurcation of interest \cite{Sirisup2005}.

\subsection{Analytical expressions for bifurcation detection}
\label{subsec:analytical}

In this section, the analytical expressions that describe the necessary statistics for bifurcation detection are summarized. Specifically, pertaining to the objective function described in Eq.~\ref{eq:bif_system_opt}, the statistics of the steady state derivative and eigenvalues at steady states are approximated. 

\subsubsection{1D Fold bifurcation}

\begin{align} \label{eq:der_dist_summary}
    E\left[\frac{\partial f_r}{\partial x}\right] &=\frac{\partial \overline{f}}{\partial x} \nonumber, \\
    Var\left(\frac{\partial f_r}{\partial x}\right) &=\left[\frac{\partial^2 {\overline{f}}}{\partial x\partial p}  \sigma^* -sgn\left( \frac{\partial \overline{f}}{\partial p}\right) \frac{\partial \sigma}{\partial x} \right]^2,
\end{align}
 
\noindent where $f_r$ is a realization of the probabilistic surrogate (indexed $r$), $\overline{f}$ is the predictive mean of the probabilistic surrogate, $\sigma$ its uncertainty and $(\sigma^*)^2$ the uncertainty of steady state parameter location (see Sec.~\ref{subsubsec_app:1d_ss_x}). All the above quantities are calculated at the steady state of the predictive mean. From Eqs.~\ref{eq:der_dist_summary} the statistics of the squared derivative can be calculated and plugged in any standard acquisition function, such as the Lower Confidence Bound (LCB):

\begin{equation} \label{eq:acq_fold_1d}
    \alpha_{LCB-FOLD-1D}(x) = E\left[\left(\frac{\partial f_r}{\partial x}\right)^2\right] - \beta Var\left(\left(\frac{\partial f_r}{\partial x}\right)^2\right),
\end{equation}
\noindent where $\beta$ is the exploration/exploitation hyperparameter.

\subsubsection{$n$-D Fold bifurcation}

\begin{align} \label{eq:lambda_fold_summary}
   E[\lambda_{r} ] &= \lambda(\mathbf{J}_{\mu}), \nonumber \\[1ex]
   Var(\lambda_r) &= \frac{1}{(v_{L_i} v_{R_i})^2} v_{L,i_1} v_{L,i_2} v_{R,j_1} v_{R,j_2} \Sigma_{J_{r_{i_1i_2j_1j_2}}},
\end{align}

\noindent where $\lambda_r$ is the eigenvalue of interest of $f_r$ at its steady state, $\mathbf{J}_\mu$ is the Jacobian of the predictive mean at its steady state, $\mathbf{v}_R, \mathbf{v}_L$ are the right and left eigenvectors of $\mathbf{J}_\mu$ respectively and $\Sigma_{J_r}$ is the covariance of elements of the Jacobian matrix. Eq.~\ref{eq:lambda_fold_summary} is written on Einstein notation. Similarly to Eq.~\ref{eq:acq_fold_1d}, the following acquisition function can be formulated:

\begin{equation} \label{eq:acq_fold_nd}
    \alpha_{LCB-FOLD-nD}(x) = E\left[\left(\lambda_{r}\right)^2\right] - \beta Var\left(\left(\lambda_{r}\right)^2\right).
\end{equation}

\subsubsection{Hopf bifurcation}

\begin{align}
   E[Re(\lambda_{r})] &= Re(\lambda(\mathbf{J}_{\mu})), \nonumber \\[1ex]
   Var(Re(\lambda_r)) &= \frac{1}{|\overline{\mathbf{v}_{L}} \mathbf{v}_{R}|^4} \mathbf{z} \mathbf{\Sigma}_{J_{r}} \mathbf{z}^\top,
\end{align}

\noindent where $\overline{\cdot}$ denotes complex conjugate and $\mathbf{z}=\mathbf{z}(\mathbf{v}_L, \mathbf{v}_R).$ (see Eq.~\ref{eq:def_z} for the definition of $\mathbf{z}$). For more details, refer to the Appendix, Sec.~\ref{sec_app:analytical}. Similarly to Eq.~\ref{eq:acq_fold_nd}, the following acquisition function can be formulated:

\begin{equation} \label{eq:acq_hopf_nd}
    \alpha_{LCB-HOPF}(x) = E\left[Re(\lambda_{r})^2\right] - \beta Var\left(Re(\lambda_{r})^2\right).
\end{equation}

\section{Results}
\label{sec:res}

\subsection{Population dynamics (1D fold bifurcation)}
\label{subsec:res1d}

\begin{figure}
    \centering
    \includegraphics[width=14.6cm, height=13.5cm]{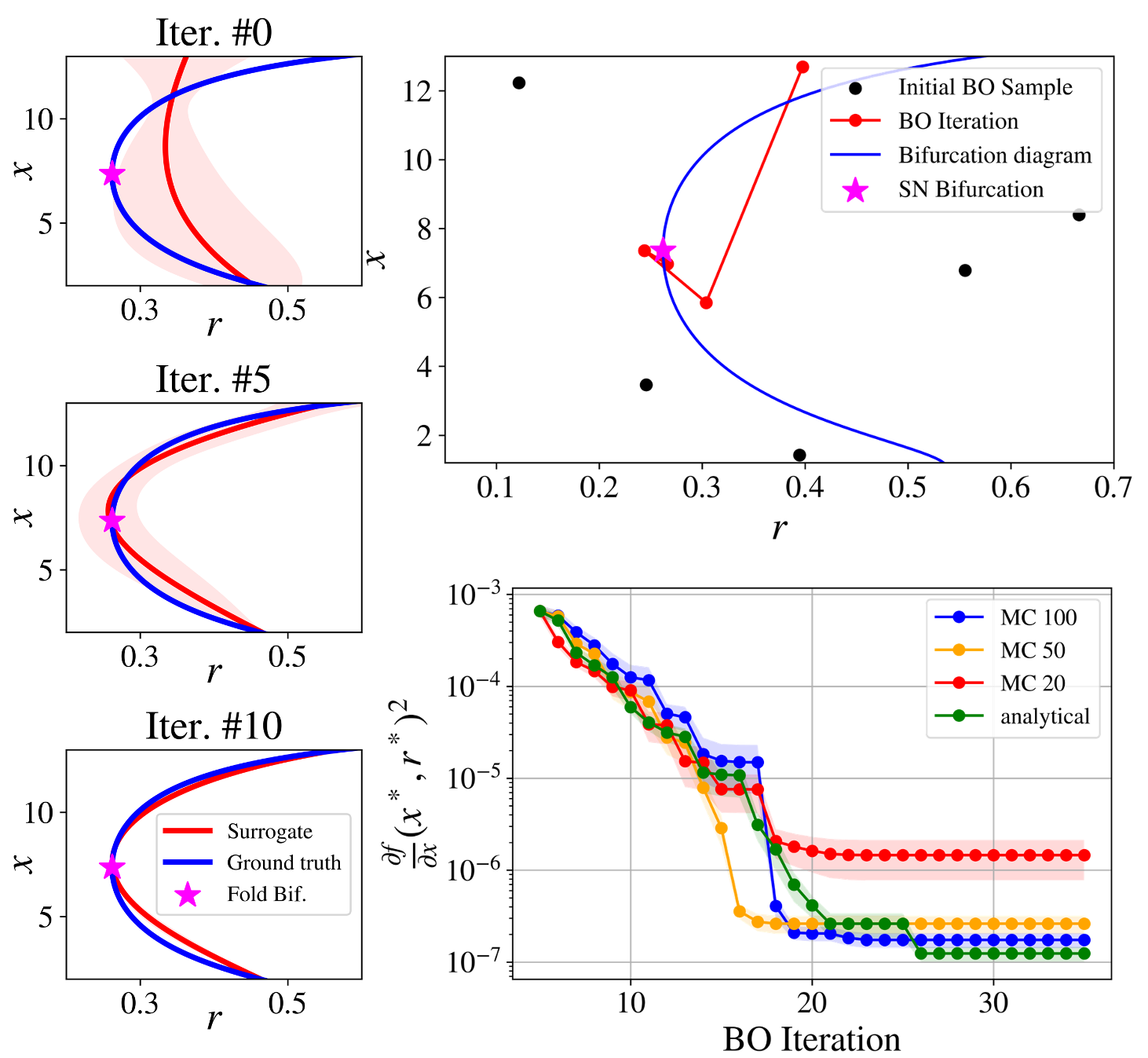}
    \caption{BO performance for the vector field in Eq.~\ref{eq:loggrowth}: On the left, the predicted bifurcation diagram (and its uncertainty) at various BO iterations is visualized overlayed on the analytical bifurcation diagram (and the exact bifurcation location), which was computed with AUTO \cite{doedel1998auto}. On the top right, a single BO trajectory is plotted on top of the bifurcation diagram, converging to the bifurcation location. On the bottom right, the performance of BO with analytical statistics is compared with BO with Monte Carlo (MC) statistics of various sample sizes for 50 BO experiments. }
    \label{fig:1d_group}
\end{figure}

The first case study is a simple, 1D nonlinear dynamical system, describing logistic growth with limited carrying capacity and death due to predation. It has been proposed to model the insect outbreak of the spruce budworm and balsam fir forest system \cite{Ludwig1978}. The vectorfield describing the dynamics is:
\begin{equation} \label{eq:loggrowth}
    \frac{dx}{dt} = f(x;r,k) \equiv  rx\left(1-\frac{x}{k}\right) -\frac{x^2}{1+x^2}, 
\end{equation}

\noindent where $x$ denotes the population (state variable), and $k,r$ are parameters. We aim to find a fold bifurcation w.r.t. $r$ for a fixed value of $k$ ($k=15$). The independent variable used to describe the solution branch is the state variable $x$ i.e. the steady state solution branch is expressed as $r(x)$. The bifurcation corresponds to a catastrophic transition between states of ``outbreak" and ``refuge".  This system serves as a simple example of catastrophe theory \cite{Zeeman1979}. 
A single experiment of our approach can be seen in the top right panel of Fig.~\ref{fig:1d_group}: BO iterates quickly converge to the (turning point, fold) bifurcation location, starting with just five vector field measurements (shown in black). The iterative updating of the Gaussian Process surrogate model translates to iterative updating of the (estimated) bifurcation diagram, as can be seen on the left of Fig.~\ref{fig:1d_group}. An ensemble of 50 BO experiments is shown in the bottom right panel of Fig.~\ref{fig:1d_group} (denoted as ``analytical"). For comparison purposes, BO performance with Monte Carlo sampling is also included, for different sampling sizes (here, 20, 50, 100). As expected, Monte Carlo sampling becomes more accurate (and less uncertain) as the sample size increases, while the analytical approach maintains high accuracy and low uncertainty with orders of magnitude lower computational cost as there is no need for Monte Carlo sampling.

\subsection{Autocatalytic reaction dynamics (2D Hopf bifurcation)}
\label{subsec:res2d_hopf}

\begin{figure}[H]
    \centering
    \includegraphics[width=10cm, height=6.67cm]{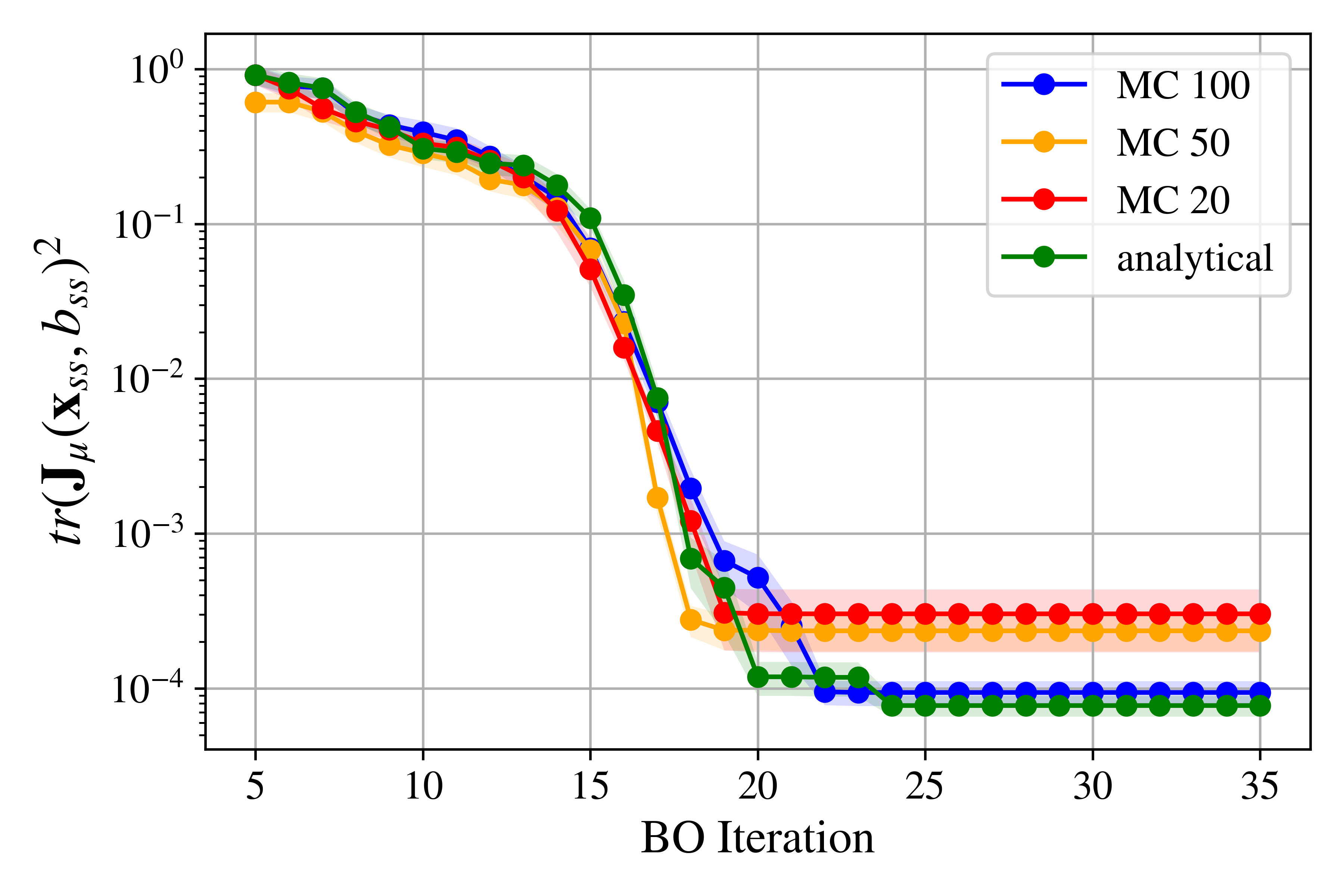}
    \caption{Bayesian Optimization convergence for the discovery of a Hopf bifurcation using the statistics of Jacobian traces. The performance of BO with analytical statistics is compared with BO with Monte Carlo statistics of various sample sizes.}
    \label{fig:2d_conv_trace}
\end{figure}

\begin{figure}[H]
    \centering
    \includegraphics[width=10cm, height=6.67cm]{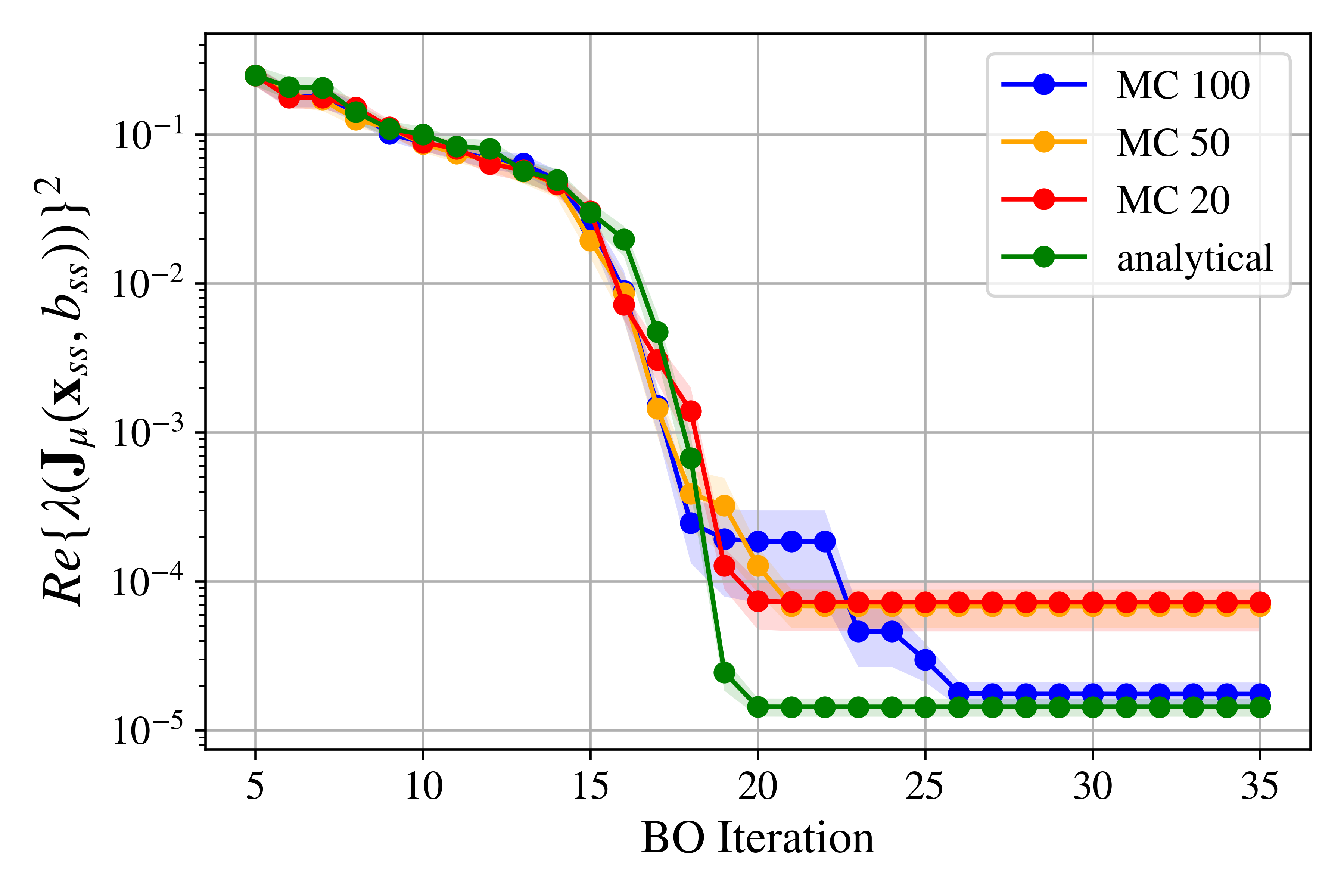}
    \caption{BO convergence for the discovery of a Hopf bifurcation using the statistics of eigenvalue real parts. The performance of BO with analytical statistics is compared with BO with Monte Carlo statistics of various sample sizes.}
    \label{fig:2d_conv_eig}
\end{figure}

The second case study is the Brusselator, a well-studied 2D benchmark for autocatalytic kinetics \cite{Prigogine1985, PhysRevE.81.046215}. The vector field describing the dynamics is:
\begin{equation} \label{eq:bru}
    \frac{d\mathbf{x}}{dt} = \mathbf{f}(\mathbf{x};a,b) \equiv
    \begin{pmatrix}
      a + x^2y - bx -x \\
      bx - x^2y\\
    \end{pmatrix},
\end{equation}

\noindent where $\mathbf{x} =(x,  y)^\top$ is the state vector and $a,b$ are parameters. The aim here is to locate a Hopf bifurcation w.r.t. the parameter $b$, for a fixed value of $a$ ($a=1.5$) marking the onset of self-sustained chemical oscillations. Such naturally occurring chemical oscillations have attracted the interest of the dynamics community since the discovery of the famous Belousov-Zhabotinsky ``clock" reaction \cite{belousov1951periodic}. In this case, we can express the solution branch as $\mathbf{x}^*(b)$. As mentioned in Sec.~\ref{subsec:formulation} for the detection of a Hopf bifurcation the criticality condition can be formulated either using the Jacobian trace or the absolute real part of the eigenvalue. For validation purposes, we use both of these criteria in Figs.~\ref{fig:2d_conv_trace} and \ref{fig:2d_conv_eig} respectively. The Monte Carlo approach is included also here for comparison purposes. As expected, the proposed BO approach manages to locate the Hopf bifurcation, while it outperforms the Monte Carlo based experiments in terms of computational efficiency.


As a demonstration of the computational benefit of the analytical approach, the associated analytical calculations take significantly less time compared even with the Monte Carlo of sample size 20. The benefits, of course are way more pronounced for higher sample sizes.

\subsection{Continuous Stirred Tank Reactor (2-D fold bifurcation)}
\label{subsec:res2d_fold}

The next case study arises in Chemical Reaction Engineering;  we study the dynamics of a Continuous Stirred Tank Reactor (CSTR), described by a system of 2 ODEs \cite{Uppal1974,  Uppal1976}: 
\begin{equation} \label{eq:cstr}
    \frac{d\mathbf{x}}{dt} = \mathbf{f}(\mathbf{x}; \mathbf{p}) \equiv
    \begin{pmatrix}
       -x_1 + Da e^{x_2}(1-x_1) \\
      -x_2 +BDae^{x_2}(1-x_1) +\beta (T_c-x_2)\\
    \end{pmatrix},
\end{equation}

\noindent where $\mathbf{x}=(x_1, x_2)^\top$ and $\mathbf{p} = (Da, B, \beta, T_c)$ are the parameters. This system exhibits a rich variety of dynamic behaviors, such as bistability and self-sustained oscillations. We choose to focus on a fold bifurcation with respect to the Damk\"ohler number $Da$. Such a bifurcation marks the onset of bistability, which  translates to two coexisting, significantly different, stable reactor operation configurations. 

Locally, the solution branch can be expressed with a state variable as the independent variable, e.g. as $x_2(x_1), Da(x_1)$. The remaining parameters are fixed: $(B, \beta, T_c) = (10, 0.1, -0.04)$.  Note that in the general case of an $n-$d fold bifurcation, a similar approach can be used, i.e. the independent variable can be chosen from elements of the state vector. 

\begin{figure}[H]
    \centering
    \includegraphics[width=10cm, height=6.67cm]{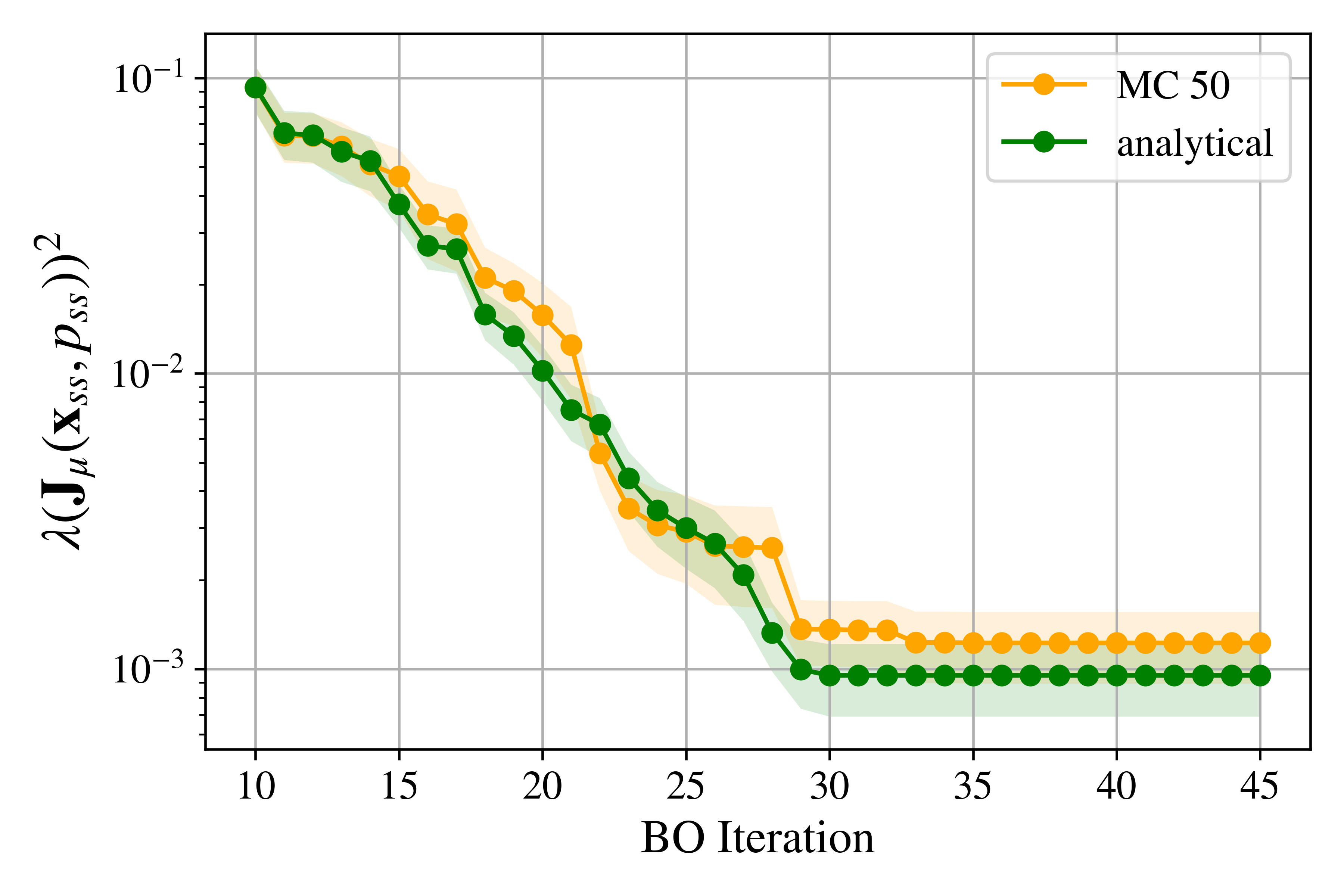}
    \caption{BO for the discovery of a fold bifurcation using the statistics of the eigenvalues' real part.}
    \label{fig:2d_fold}
\end{figure}

BO convergence results (over an ensemble of 50 experiments) can be seen in Fig.~\ref{fig:2d_fold}. BO manages to approximate the bifurcation point, and outperforms the Monte Carlo alternative with 50 samples.

\textcolor{red}{}

\subsection{Epileptor - neuronal dynamics (4D fold bifurcation)}
\label{subsec:res4d}

\begin{figure}[H]
    \centering
    \includegraphics[width=12.1cm, height=11cm]{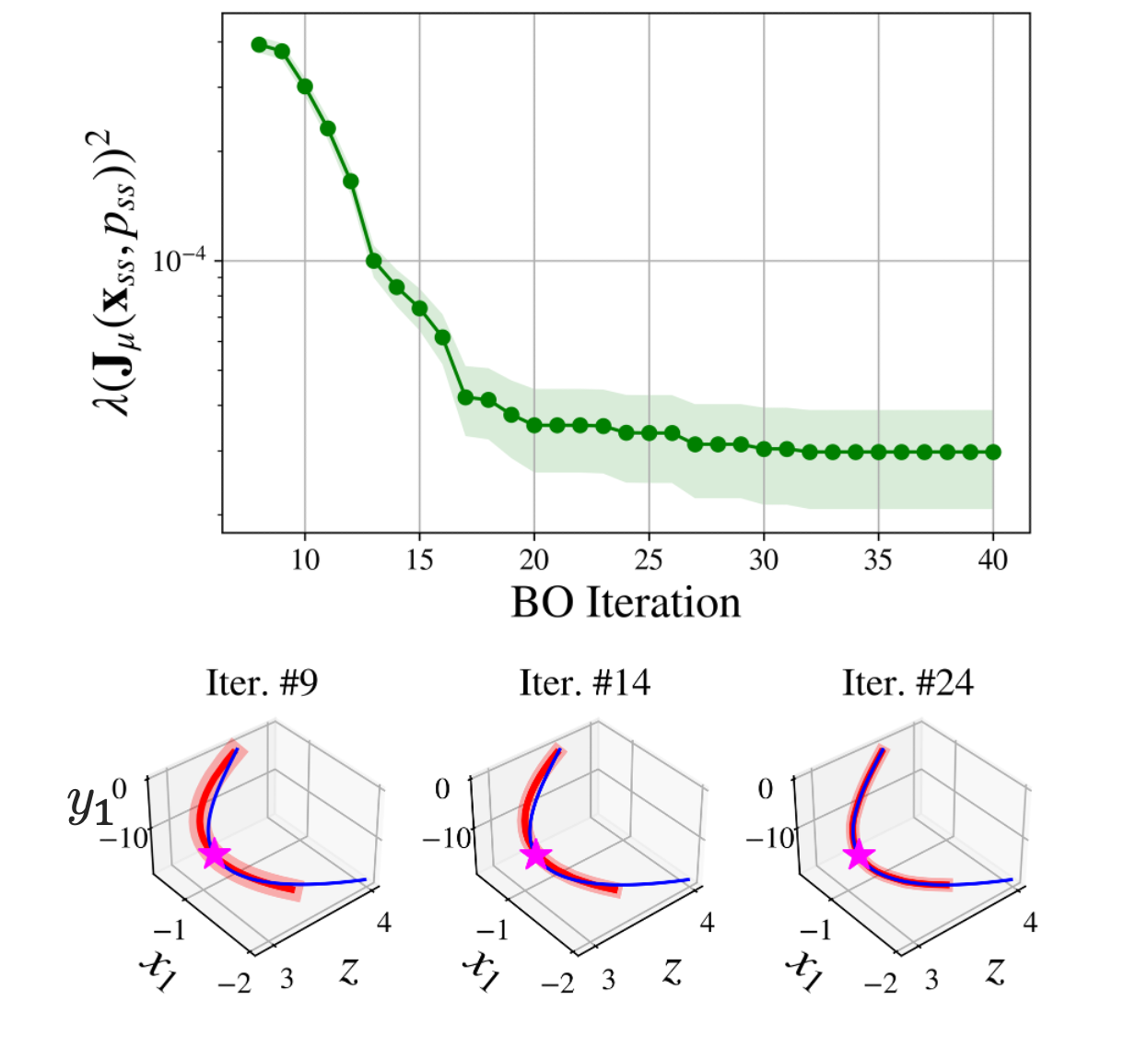}
    \caption{(top) Bayesian Optimization convergence for the discovery of a fold bifurcation for the Epileptor model in Sec.~\ref{subsec:res4d}. Representative uncertainty of all BO experiments is also shown (semi-transparent green intervals). (bottom) Updating of the bifurcation diagram  for the same model along the BO iterates (representative uncertainty also shown).}
    \label{fig:epi_conv}
\end{figure}

The next case study presented is the epileptor, a 4D dynamical system from the field of computational neuroscience, describing seizure dynamics \cite{ElHoussaini2020, Zhang2018}: 
\begin{equation} \label{eq:epi}
    \frac{d\mathbf{x}}{dt} = \mathbf{f}(\mathbf{x}; \mathbf{p}) \equiv
    \begin{pmatrix}
       y_1 -f_1(x_1, x_2)-z+I_{ext_1} \\
       c_1-d_1x_1^2-y_1\\
       -y_2+x_2-x_2^3+I_{ext_2}-0.3(z-3.5)\\
       \frac{f_2(x_2)-y_2}{\tau_2}\\
    \end{pmatrix},
\end{equation}

\[ 
\text{where} \hspace{0.2cm} f_1(x_1,x_2)= \left\{
\begin{array}{ll}
      & ax_1^3-bx_1^2, \hspace{0.2cm} x_1<0  \\ &  -(m-x_2+0.6(z-4)^4)x_1, \hspace{0.2cm} x_1\geq 0,\\
\end{array} 
\right. 
\]
\[ 
\text{and} \hspace{0.2cm}f_2(x_2)= \left\{
\begin{array}{ll}
      & 0, \hspace{0.2cm} x_2<-0.25  \\ &  a_2(x_2+0.25) \hspace{0.2cm} x_2\geq -0.25,\\
\end{array} 
\right. 
\]
\noindent where $\mathbf{x}=(x_1, y_1, x_2, y_2)^\top$ is the state vector and $\mathbf{p} = (z, I_{ext_1}, c_1, d_1, I_{ext_2}, \tau_2, a, b, m, a_2) $ the parameter vector. When the parameter $z$ slowly varies, it may drive the system towards a tipping point (fold bifurcation \cite{ElHoussaini2020}) which marks the onset of epileptic seizures (in relevant literature: transition from normal to ictal (i.e. critical) states \cite{Jirsa2014}). Therefore, it becomes important to study bifurcations with respect to the parameter $z$. The rest of the parameters are fixed: $(I_{ext_1}, c_1, d_1, I_{ext_2}, \tau_2, a, b, m, a_2)  = (3.1, 1, 5, 0.45, 10, 1 , 3, 0.5, 6)$. As in Sec.~\ref{subsec:res2d_fold} we choose a state variable as the independent variable with respect to which we express the solution branch locally. Here $x_1$ is chosen as the independent variable. In Fig.~\ref{fig:epi_conv} the convergence of a sample of 50 Bayesian Optimization experiments is shown, using the analytical statistics formulas (see \ref{sec_app:analytical} in the Appendix). Again, the proposed BO methodology manages to locate the fold bifurcation, even for this higher-dimensional system.

\subsection{FitzHugh-Nagumo PDE}
\label{subsec:res_pde}

The last case study to be discussed is the FitzHugh-Nagumo PDE,  a reaction-diffusion PDE describing the time evolution of two coupled fields $u(x,t),v(x,t) $ in a single spatial dimension $x$ \cite{Anderson1999}. This system has been extensively used as a prototype of an activator-inhibitor dynamical system \cite{Lee2020, OWilliams2015}:
\begin{align}
    u_t =& D_uu_{xx} +u -u^3 -v \nonumber \\
    v_t =& D_v v_{xx}+\epsilon(u-\alpha_1v-\alpha_0), \label{eq:fhn_pde}
\end{align}

\noindent with $u=u(x,t)$, $v=v(x,t)$, $x\in[0,L]$,  homogeneous von Neumann boundary conditions and parameter vector $\mathbf{p}=(D_u, D_v,\alpha_1, \alpha_0, \epsilon)$.  Steady states (and bifurcations) of such a PDE can be found numerically, e.g. via a finite differences approximation. Here a second-order central finite differences scheme was used with $\Delta x = 0.1$ and $L=20$. \cite{Galaris2022}. This set of PDEs is known to exhibit a fold bifurcation (among others), with respect to parameter $\epsilon$, when all other parameters are fixed at $(D_u, D_v,\alpha_1, \alpha_0)=(1,4,2, -0.03)$ \cite{Galaris2022}. 

The result of a finite differences approximation is a large number of coupled ODEs (specifically, with the discretization described above, 402 ODEs). Assuming that the underlying dynamics are really lower-dimensional we explore reduced order descriptions via Proper Orthogonal Decomposition (POD)
\cite{Deane1991, GEAR2002941, Sirisup2005}.
Here, the POD basis was constructed based on 10 empirically chosen state vectors, representative of the bifurcation diagram. Eventually, 8 POD components were retained as sufficient to describe the PDE dynamics.
However, as mentioned in \ref{subsec:abm}, it is often possible that the last few POD components are functions of the leading few \cite{koronaki2023nonlinear}. This is exploited here, to further reduce dimensionality: the last four POD components are described as a function of the leading four, resulting in a 4-dimensional reduced order model for the Fithugh-Nagumo PDE.

\begin{figure}[H]
    \centering
    \includegraphics[width=10cm, height=6.67cm]{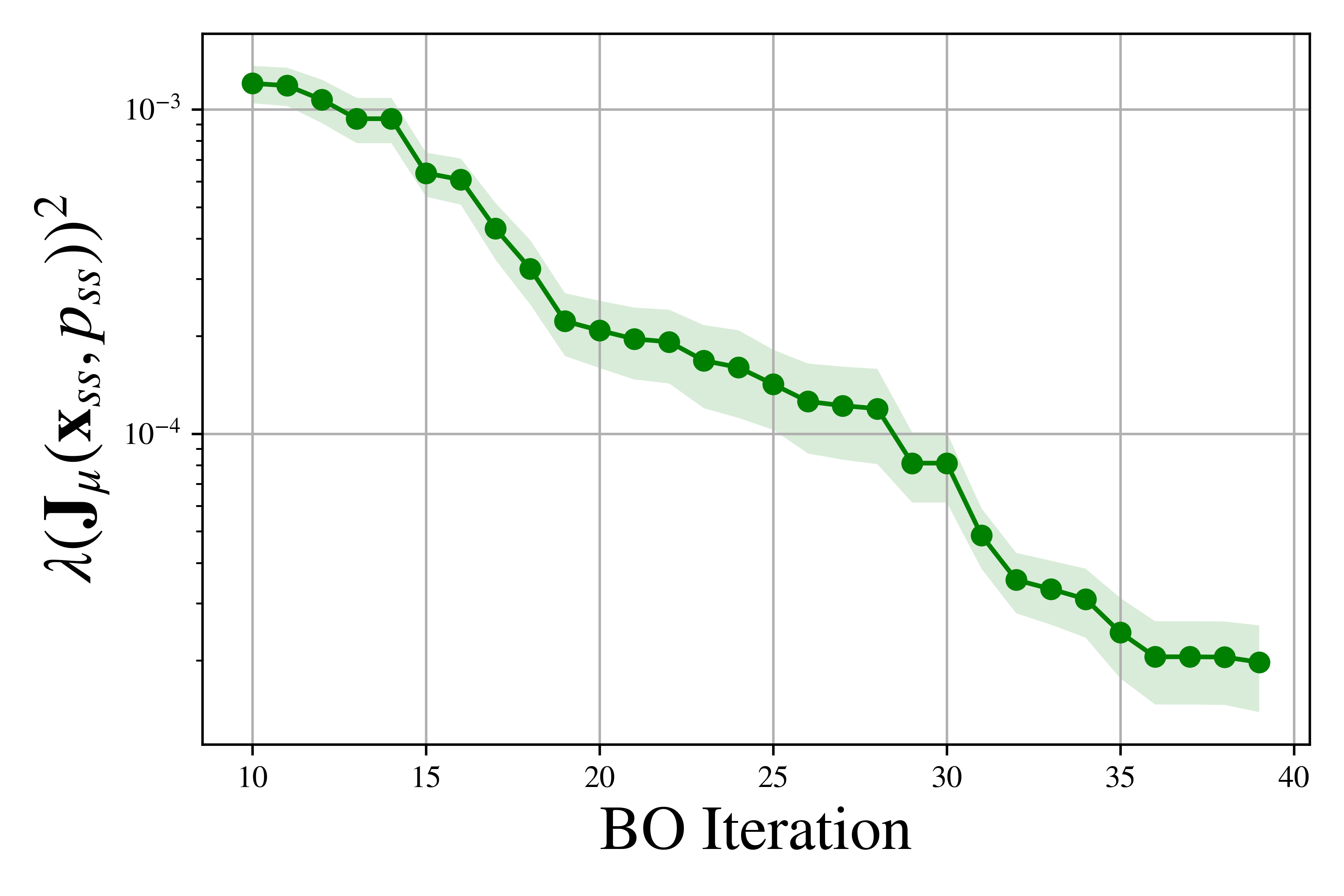}
    \caption{Bayesian Optimization convergence for the discovery of a fold bifurcation for the POD-reduced FitzHugh-Nagumo PDEs (Sec.~\ref{subsec:res_pde}).}
    \label{fig:fhn_pde}
\end{figure}

Given that the bifurcation of interest is a fold bifurcation, an independent variable is chosen out of this reduced state vector. Here, the second POD component ($a_2$) is selected. The convergence result of a sample of 30 BO experiments is shown in Fig.~\ref{fig:fhn_pde}. Evidently, the proposed BO active search protocol locates the fold bifurcation of the FitzHugh-Nagumo PDE.

\section{Conclusions}
\label{sec:conc}

We have formulated a Bayesian optimization framework for the detection of bifurcations in systems for which it is hard to perform a large number of calculations or experiments. The key ingredient of the proposed formulation is the derivation of analytical expressions, and their uncertainty, for quantities related to bifurcation detection, e.g. derivatives and eigenvalues at steady states. We are able to derive such expressions by relying on a Gaussian Process regression scheme for the right hand side of the unknown dynamical system, and the use of asymptotic analysis with respect to the underlying variance. 

Using the developed active learning algorithm, we showed how bifurcations can be discovered in simple dynamical systems, such as population growth dynamics, chemical kinetics, reactor dynamics and neuronal dynamics. Even if these models were selected as proof of concept, they illustrate the algorithm's potential: discovery of ecological catastrophes, emergence of chemical clocks, dramatic reactor operation transitions and initiation of epileptic shocks. Note that such bifurcations possibly translate to rare events - tipping points in stochastic contexts \cite{Siettos_2012}.

More specifically, the results in Sec.~\ref{sec:res} illustrate the ability of active learning --and specifically, Bayesian Optimization-- to efficiently locate fold or Hopf bifurcations. To validate our approach we compared the developed BO framework that uses analytical expressions for the eigevanlues or derivatives to a Monte Carlo alternative. As expected, the developed method outperforms Monte Carlo estimates and manages to locate bifurcations more efficiently and accurately, with orders of magnitude lower computational cost. In addition, the active learning character of the method allows for detection of bifurcations in resource-constrained problems, since only a handful of rudimentary observations are necessary to get good convergence.

We emphasize that the developed algorithms can be easily adjusted to handle multiple coexisting bifurcations, bifurcations of higher codimension (e.g. cusp bifurcations) and  bifurcations of periodic orbits. Moroever, the work showcases how uncertainty quantification can be utilized when performing standard numerical tasks for dynamical systems: uncertainty in steady state location, eigenvalue estimation and discovery of bifurcations. This is in agreement with previous work of the coauthors where Bayesian Continuation \cite{Psarellis2023} was introduced. 

Looking into the future, cross-fertilization between numerical methods and uncertainty quantification holds great promise; numerical errors in tasks such as root-finding, integration \cite{Tronarp2021, schober2014probabilistic} or continuation can be thought as variance of a random variable \cite{Hennig2022}. A natural consequence is that scientific computing algorithms that are adaptive to error, could be recast as \textit{adaptive to uncertainty} \cite{Hennig2022}. This change in perspective allows the utilization of advanced uncertainty quantification tools and algorithms, and, in turn, Machine Learning, to perform adaptive scientific computing. Hopefully, the present work serves as an example of this cross fertilization, specific for dynamical systems: standard analysis of a (hidden) dynamical system performed through a machine learning surrogate, and search strategies developed for active learning, applied to the location of bifurcations.

\subsection*{Acknowledgments} TPS acknowledges support from AFOSR (MURI grant
FA9550-21-1-0058) as well as ONR (grant N00014-21-1-2357). The work of YMP and IGK was partially supported
through an ARO MURI (via UCSB) and the US Air Force Office of Scientific Research. 
\newpage

\section*{Appendix}
\appendix

\section{Analytical Uncertainty Calculations}
\label{sec_app:analytical}

We explore how uncertainty propagates from vectorfield values to derivatives (or eigenvalues of the linearization) obtained at the vectorfield steady state(s).
Specifically, starting from (possibly noisy, or sparse) observations of the vectorfield of a parameter-dependent dynamical system, we first train a surrogate model for the vectorfield (via Gaussian Process in this case). Following that, we estimate the distribution of  steady states. Note, that such calculations are local, in the neighborhood of either for a fixed parameter or a fixed state. Finally, we estimate the distribution of the corresponding derivatives (or eigenvalues) of the vectorfield at such steady states.\\ In the following, we will be using the general notation:

\begin{itemize}
    \renewcommand{\labelitemi}{\scriptsize$\blacksquare$}
    \item $\mathbf{f_{vf}}(\mathbf{x}, p):\mathbb{R}^{n+1} \rightarrow \mathbb{R}^n $: the unknown vector field ($\dot{\mathbf{x}} = \mathbf{f_{vf}}(\mathbf{x}, p) $), where $\mathbf{x} \in \mathbb{R}^n$  is the state vector  and $p \in \mathbb{R}$ is a scalar parameter.
    \item $\overline{\mathbf{f}}(\mathbf{x}, p):\mathbb{R}^{n+1} \rightarrow \mathbb{R}^n $: predictive mean of the GP surrogate model.
    \item  $\mathbf{\Sigma}(\mathbf{x}, p):\mathbb{R}^{n+1} \rightarrow \mathbb{R}^{n \times n} $: covariance matrix (uncertainty) as calculated in the GP framework (this reduces to $\sigma^2$(x, p) in 1D). Each component corresponds to $\mathrm{Cov}(f_i, f_j)$.
    \item $\mathbf{f}_r(\mathbf{x}, p):\mathbb{R}^{n+1} \rightarrow \mathbb{R}^n $: A random GP realization, indexed by $r$.
\end{itemize}
 
\noindent We will consider the following cases:

\begin{enumerate}
    \item 1-D problem, for a fold bifurcation detection (Sec.~\ref{subsec_app:1d}).
    \item $n-$D problem, for a Hopf bifurcation detection (Sec.~\ref{subsec_app:hopf}).
\end{enumerate}

\noindent \textbf{Our main assumption is that $\mathbf{\Sigma (x}, p)$  (or, $\sigma(x,p) $ in 1-D) has small values.}

\subsection{1-D, fold bifurcation detection}
\label{subsec_app:1d}

In this section, we deal with the case where the state is one dimensional ($n=1$). In order to find steady state distributions, we will either fix the parameter $p$ or the state $x$. Subsequently, distributions of derivatives at such steady states are estimated (Sec.~\ref{subsubsec_app:derfold}). For the purposes of detecting fold bifurcations,  only the case of fixed $x$ is explored, as the implicit function theorem restricts us to expressing the solution branch as $p(x)$. The statistics of the \textit{squared} derivatives are also derived (in Sec.~\ref{subsubsec_app:sqderfold}) which are useful in the context of minimization.

\subsubsection{Steady state distribution, fixing $x$}
\label{subsubsec_app:1d_ss_x}

Here, the state $x$ has a fixed value, which implies a (1D) distribution in parameter space for the steady states (to be precise, for the the steady state parameters).
Note that this approach implies that the solution branch is expressed as $p(x)$ and is suitable for the detection of fold bifurcations.\\

\noindent Notation relevant to this subsection:

\begin{itemize}
    \renewcommand{\labelitemi}{\scriptsize$\blacksquare$}
    \item $p_\mu^*\in \mathbb{R}$: the root of the predictive mean ($\overline{f}(x,p^*_\mu)=0$) given a fixed value of $x$.
    \item $p_r^* \in \mathbb{R}$: the root of the GP realization indexed by $r$ ($f_r(x,p^*_r)=0$)  given a fixed value of $x$.
    \item $(\sigma^*)^2 \equiv Var(p_r^*)$: the variance (uncertainty) of the steady states (steady state parameters), given a fixed value of $x$.
\end{itemize}

\begin{align}
& \textcolor{black}{\text{For every GP realization indexed by $r$ there is a unique $\theta_r \sim \mathcal{N}(0,1)$ such that:}} \notag \\[1ex]
& f_r(x, p_r^*) = \overline{f}(x, p_r^*)+\theta_{r}\sigma(x, p_r^*). \label{eq:al3_1} \\[1ex]
%
%
& \textcolor{black}{\text{By definition, $f_r(x, p_r^*)=0$. Expressing the steady state of realization $r$ as $p^*_r = p_\mu^*+ \rho_{r}^*$},} \notag  \\
& \textcolor{black}{\text{$\rho_{r}^* \sim \mathcal{N}(0,(\sigma^*)^2)$  and substituting in Eq.~\ref{eq:al3_1}:}}\notag  \\[1ex]
%
%
&\overline{f}( x,p_\mu^*+ \rho_{r}^*) + \theta_{r}\sigma( x, p_\mu^*+ \rho_{r}^*)=0. \label{eq:8}\\[1ex]
& \textcolor{black}{\text{Using a Taylor expansion for functions $\overline{f}(x,\cdot), \sigma(x,\cdot)$ around $p_\mu^*$:}}  \notag \\[1ex] 
%
%
&\overline{f}( x, p_\mu^*)+\rho_{r}^*\frac{\partial \overline{f}}{\partial p}( x, p_\mu^*)+ \theta_{r}\sigma(x, p_\mu^*)+ \theta_{r}\rho_{r}^*\frac{\partial \sigma}{\partial p}(x, p_\mu^*)= \mathcal{O}( (\rho_{r}^*)^2). \\[1ex]
%
%
%
%
%
%
&\textcolor{black}{\text{Assuming $|\rho_r^*|$ is small, we can omit the remainder $\mathcal{O}((\rho_r^*)^2)$ and set $\rho_{r}^*\frac{\partial \sigma}{\partial p}(x, p_\mu^*)  \approx 0$: }} \notag \\[1ex]
%
%
&\overline{f}( x, p_\mu^*)+\rho_r^*\frac{\partial\overline{f}}{\partial p}( x, p_\mu^*)+ \theta_{r}\sigma(x, p_\mu^*) \approx 0. \label{eq:full20}\\[1ex]
& \textcolor{black}{\text{By definition $\overline{f}( x, p_\mu^*)=0$. Then:}}  \notag \\[1ex]
&Var(\rho_r^*) = (\sigma^*)^2 = \left(\frac{\partial \overline{f}}{\partial p}(x, p_\mu^*)\right)^{-2}\sigma^2(x, p_\mu^*).
\end{align}
\noindent Therefore the distribution of the  steady states will be:
\begin{equation} \label{eq:71}
    (x, p^*_r)\hspace{0.2cm} \text{where, } \hspace{0.2cm} p^*_r \sim \mathcal{N}\left(p_\mu^*,   \left(\frac{\partial \overline{f}}{\partial p}(x, p_\mu^*)\right)^{-2}\sigma^2(x, p_\mu^*) \right).
\end{equation}

\subsubsection{Derivative distribution for fold detection, fixing $x$}
\label{subsubsec_app:derfold}

In this section, the distribution of the state derivatives at the steady states is estimated. 
For the purpose of creating an active learning protocol for fold bifurcation detection, we will choose to fix $x$ (as in Sec.~\ref{subsubsec_app:1d_ss_x}) and the steady states will follow the distribution:
\begin{equation} \nonumber
    (x, p^*_r)\hspace{0.2cm} \text{where, } \hspace{0.2cm} p^*_r \sim \mathcal{N}\left(p_\mu^*,   \left(\frac{\partial \overline{f}}{\partial p}(x, p_\mu^*)\right)^{-2}\sigma^2(x, p_\mu^*) \right),
\end{equation}

\begin{align}   
& \textcolor{black}{\text{Using Eq.~\ref{eq:al3_1} as a starting point, the state derivatives (w.r.t. $x$) at the steady states can be expressed as:}} \notag \\[1ex]
&\frac{\partial f_r}{\partial x}(x,p^*_r) = 
\frac{\partial \overline{f}}{\partial x}(x, p_r^*) + \theta_{r}\frac{ \partial \sigma}{\partial x}( x,p_r^*). \\[1ex]
%
%
&\textcolor{black}{\text{Substituting $p_r^*=p_\mu^*+\rho_{r}^*$, as in Eq.~\ref{eq:8}:}}\notag \\[1ex]
%
%
&\frac{\partial f_r}{\partial x}(x,p^*_r) =  
\frac{\partial \overline{f}}{\partial x}(x, p_\mu^*+ \rho_{r}^*) + \theta_{r}\frac{ \partial \sigma}{\partial x}( x,p_\mu^*+ \rho_{r}^*). \\[1ex]
& \textcolor{black}{\text{Expanding the functions $\frac{\partial \overline{f}}{\partial x}(x, \cdot), \frac{\partial \sigma}{\partial x}(x, \cdot)$ using a Taylor series:}} \notag \\[1ex]
%
%
&\frac{\partial f_r}{\partial x}(x,p^*_r) =  \frac{\partial \overline{f}}{\partial x}(x, p_\mu^*) + \rho_{r}^*\frac{\partial^2 \overline{f}}{\partial x \partial p}(x, p_\mu^*) + 
\theta_r\frac{\partial \sigma}{\partial x}(x, p_\mu^*) + \theta_r\rho_{r}^*\frac{\partial^2 \sigma}{\partial x \partial p}(x, p_\mu^*) + \mathcal{O}((\rho_{r}^*)^2).  \\[1ex]
%
%
%
%
%
& \textcolor{black}{\text{Assuming again that $|\rho_r^*|$ is small, we can eliminate the remainder $\mathcal{O}((\rho_r^*)^2)$ and set $\rho_{r}^*\frac{\partial^2 \sigma}{\partial x \partial p}(x, p_\mu^*) \approx 0$ :}} \notag  \\[1ex]
%
%
&\frac{\partial f_r}{\partial x}(x,p^*_r) \approx  \frac{\partial \overline{f}}{\partial x}(x, p_\mu^*) +\rho_r^*\frac{\partial^2 \overline{f}}{\partial x \partial p}(x, p_\mu^*) +  
\theta_r\frac{\partial \sigma}{\partial x}(x, p_\mu^*). \\[1ex]
&\textcolor{black}{\text{Using Eq.~\ref{eq:full20}, we can express $\rho_{r}^* = -sgn\left( \frac{\partial \overline{f}}{\partial p}(x,p_\mu^*)\right)\sigma^* \theta_r$: }} \notag \\[1ex]
&\frac{\partial f_r}{\partial x}(x,p^*_r) \approx  \frac{\partial \overline{f}}{\partial x}(x, p_\mu^*) -sgn\left( \frac{\partial \overline{f}}{\partial p}(x,p_\mu^*)\right) \theta_r \sigma^*\frac{\partial^2 \overline{f}}{\partial x \partial p}(x, p_\mu^*) +  
\theta_r\frac{\partial \sigma}{\partial x}(x, p_\mu^*). \\[1ex]
& \textcolor{black}{\text{Then:}} \notag \\[1ex]
&E\left[\frac{\partial f_r}{\partial x}(x,p^*_r)\right] = \frac{\partial \overline{f}}{\partial x}(x,p^*_\mu), \label{eq:mean_der}\\[1ex]
&Var\left(\frac{\partial f_r}{\partial x}(x,p^*_r)\right)=\left[\frac{\partial^2 {\overline{f}}}{\partial x\partial p}(x,p^*_\mu) \sigma^* -sgn\left( \frac{\partial \overline{f}}{\partial p}(x,p_\mu^*)\right) \frac{\partial \sigma}{\partial x}(x,p^*_\mu) \right]^2. \label{eq:var_der} 
\end{align}

Therefore, the distribution of the state derivatives at the steady states can be approximated as:

\begin{equation} \label{eq:der_dist}
    \frac{\partial f_r}{\partial x}(x,p^*_r) \sim \mathcal{N}\left(\frac{\partial \overline{f}}{\partial x}(x,p^*_\mu),\left[\frac{\partial^2 {\overline{f}}}{\partial x\partial p}(x,p^*_\mu)  \sigma^* -sgn\left( \frac{\partial \overline{f}}{\partial p}(x,p_\mu^*)\right) \frac{\partial \sigma}{\partial x}(x,p^*_\mu) \right]^2\right), 
\end{equation}
 
\noindent where $(\sigma^*)^2=\left(\frac{\partial \overline{f}}{\partial p}(x,p_\mu^*)\right)^{-2}\sigma^2(x,p_\mu^*)$ (see Sec.~\ref{subsubsec_app:1d_ss_x}).

\subsubsection{\textit{Squared} derivative distribution for fold detection, fixing $x$}
\label{subsubsec_app:sqderfold}

To design an active learning protocol (here, Bayesian Optimization) for fold bifurcation detection, it is necessary to formulate an objective function which assumes an optimum at the location of the fold bifurcation. One option is to minimize the squared state derivative at the steady state for each fixed $x$ value.

Then, we would need the expected value and the variance of the \textit{\textbf{squared}} derivatives. To calculate them, we can use the noncentral Gaussian moments (see property ~\ref{prop:1}) to obtain:\\ 

\begin{align}
&E\left[\frac{\partial f_r}{\partial x}(x,p^*_r)^2\right] = E\left[\frac{\partial f_r}{\partial x}(x,p^*_r)\right]^2 + Var\left(\frac{\partial f_r}{\partial x}(x,p^*_r)\right), \\[1ex]
&Var\left(\frac{\partial f_r}{\partial x}(x,p^*_r)^2\right) = 2Var\left(\frac{\partial f_r}{\partial x}(x,p^*_r)\right) \left(2 E\left[\frac{\partial f_r}{\partial x}(x,p^*_r)\right]^2 + Var\left(\frac{\partial f_r}{\partial x}(x,p^*_r)\right)  \right),
\end{align}

where $E\left[\frac{\partial f_r}{\partial x}(x,p^*_r)\right], Var\left(\frac{\partial f_r}{\partial x}(x,p^*_r)\right)$ can be estimated through Eqs.~\ref{eq:mean_der}, \ref{eq:var_der}.\\

\subsection{$n$-D, Hopf bifurcation detection}
\label{subsec_app:hopf}

In this section, the state space is $n-$dimensional. We can assume that in the vicinity of a Hopf bifurcation (which we are interested in), the solution branch can be expressed as $\mathbf{x}(p)$.  Analogously with the 1-D case of Sec.~\ref{subsubsec_app:1d_ss_x}, in the following (Sec.~\ref{subsubsec_app:hopf_ss}) we will first derive the distribution of steady states, now fixing the parameter $p$. Then, in Sec.~\ref{subsubsec_app:eighopf}, the distribution of Jacobians at steady states is estimated, followed by the distribution of the respective eigenvalues (Sec.~\ref{subsubsec_app:eigv}) and the respective squared eigenvalues.

\subsubsection{Steady state distribution, fixing $p$}
\label{subsubsec_app:hopf_ss}

In this subsection, the ($n-$dimensional) distribution of steady states is estimated, for a fixed parameter $p$.
\noindent Notation relevant to this subsection:

\begin{itemize}
    \renewcommand{\labelitemi}{\scriptsize$\blacksquare$}
    \item $\mathbf{x}_\mu^*\in \mathbb{R}^n$: the root of the predictive mean ($\overline{\mathbf{f}}(\mathbf{x}^*_\mu,p)=\mathbf{0}$), given a fixed value of $p$.
    \item $\mathbf{J}_\mu(\mathbf{x},p): \mathbb{R}^{n+1} \rightarrow \mathbb{R}^{n\times n}$: the Jacobian of the predictive mean ($\mathbf{J}_\mu(\mathbf{x},p) \equiv \frac{\partial \overline{\mathbf{f}}}{\partial \mathbf{x}}(\mathbf{x}, p)$)
    \item $\mathbf{x}_r^* \in \mathbb{R}^n$: the root of the GP realization indexed by $r$ ($\mathbf{f}_r(\mathbf{x}^*_r,p)=\mathbf{0}$),  given a fixed value of $p$.
    \item $\mathbf{J}_r(\mathbf{x},p): \mathbb{R}^{n+1} \rightarrow \mathbb{R}^{n\times n}$: the Jacobian of the GP realization indexed by $r$ ($\mathbf{J}_r(\mathbf{x},p) \equiv \frac{\partial \mathbf{f_r}}{\partial \mathbf{x}}(\mathbf{x}, p)$)
    \item $\mathbf{\Sigma^*} \equiv \mathrm{Cov}(\mathbf{x}_r^*) \in \mathbb{R}^{n \times n}$: the covariance matrix (uncertainty) of the steady states,  given a fixed value of $p$.
    \item $\mathbf{B}^* \in \mathbb{R}^{n \times n}$: The Cholesky decomposition of $\mathbf{\Sigma}^*$ ($\mathbf{\Sigma^*} = \mathbf{B^*} \mathbf{B^*}^\top$),  given a fixed value of $p$.
    \item $\mathbf{B}(\mathbf{x},p): \mathbb{R}^{n+1} \rightarrow \mathbb{R}^{n \times n}$: The Cholesky decomposition of $\mathbf{\Sigma}(\mathbf{x},p)$ ($\mathbf{\Sigma}(\mathbf{x},p) = \mathbf{B}(\mathbf{x},p) \mathbf{B}(\mathbf{x},p)^\top$).
\end{itemize}

In the following, equations will be written in Einstein notation.

\begin{align}
& \textcolor{black}{\text{For every GP realization indexed by $r$ there is a unique $\mathbf{\Theta}_r \sim \mathcal{N}(\mathbf{0},\mathbf{I}_n)$ (where $\mathbf{I}_n \in \mathbb{R}^{n\times n}$}} \notag \\
&\textcolor{black}{\text{is the identity matrix), such that:}} \notag \\[1ex]
%
%
%
&\textcolor{black}{f_{r_i}(\mathbf{x}_r^*,p) = \overline{f}_i(\mathbf{x}_r^*,p)+B_{ij}(\mathbf{x}_r^*,p)\Theta_{r_j}}. \label{eq:f_ndim}\\[1ex]
%
%
& \textcolor{black}{\text{By definition, $f_{r_i}(\mathbf{x}_r^*,p)=0$. Substituting $\mathbf{x}^*_r = \mathbf{x}_\mu^*+ \mathbf{P}_{r}^*, \hspace{0.2cm} \mathbf{P}_{r}^* \sim \mathcal{N}(\mathbf{0}, \mathbf{\Sigma^*})$}: } \notag \\[1ex]
%
%
%
&\textcolor{black}{\overline{f}_i( \mathbf{x}_\mu^*+ \mathbf{P}_{r}^*,p) + B_{ij}(  \mathbf{x}_\mu^*+ \mathbf{P}_{r}^*,p)\Theta_{r_j}=0}. \\[1ex]
& \textcolor{black}{\text{Expanding the functions $\overline{f}_i(\cdot, p), B_{ij}(\cdot, p)$ around  $\mathbf{x}_\mu^*$ using Taylor series:}}  \notag \\[1ex] 
%
%
%
& \textcolor{black}{\overline{f}_i( \mathbf{x}_\mu^*,p)+\frac{\partial \overline{f}_i}{\partial x_j}( \mathbf{x}_\mu^*,p)P_{r_j}^*+ 
B_{ij}(\mathbf{x}_\mu^*,p)\Theta_{r_j} + \frac{\partial B_{ij}}{\partial x_k}
(\mathbf{x}_\mu^*,p) P^*_{r_k}\Theta_{r_j}= \mathcal{O}(||\mathbf{P}_{r}^*||^2_2)}. \\[1ex]
%
%
&\textcolor{black}{\text{Assuming} \hspace{0.1cm} ||\mathbf{P}_{r}^*||_2 \hspace{0.1cm} \text{is small, we can eliminate the remainder $\mathcal{O}(||\mathbf{P}_{r}^*||^2_2)$}} \notag \\
& \textcolor{black}{\text{and the term containing products of elements of $ \mathbf{P}_{r}^*, \frac{\partial \mathbf{B}}{\partial \mathbf{x}}(\mathbf{x}_\mu^*,p)$: }} \notag \\[1ex]
%
%
%
&\textcolor{black}{\overline{f}_i( \mathbf{x}_\mu^*,p)+\frac{\partial \overline{f}_i}{\partial x_j}( \mathbf{x}_\mu^*,p)\mathbf{\Rho}_{r_j}^*+ 
B_{ij}(\mathbf{x}_\mu^*,p)\Theta_{r_j}  \approx 0}.
\\[1ex]
& \textcolor{black}{\text{By definition $\overline{f}_i( \mathbf{x}_\mu^*,p)=0$, so:}} \notag \\[1ex]
%
%
%
%
&\mathbf{\Sigma}^*  = \mathrm{Cov}(\mathbf{\Rho}_{r}^*)  
= \left(\mathbf{J_\mu}( \mathbf{x}_\mu^*,p)^{-1}\mathbf{B}(\mathbf{x}_\mu^*,p) \right)\left( \mathbf{J_\mu}( \mathbf{x}_\mu^*,p)^{-1}\mathbf{B}(\mathbf{x}_\mu^*,p)  \right)^\top  \notag \\[1ex]
&\hspace{2.2cm} = \mathbf{J_\mu}^{-1}( \mathbf{x}_\mu^*,p)\mathbf{B}(\mathbf{x}_\mu^*,p) \mathbf{B}(\mathbf{x}_\mu^*,p)^\top(\mathbf{J_\mu}^{-1} ( \mathbf{x}_\mu^*,p))^\top \notag \\[1ex]
&\hspace{2.2cm}  = \mathbf{J_\mu}^{-1}( \mathbf{x}_\mu^*,p)\mathbf{\Sigma}(\mathbf{x}_\mu^*,p)\mathbf{J_\mu}^{-1} ( \mathbf{x}_\mu^*,p)^\top .
\end{align}

\noindent Therefore the steady states will be $(\mathbf{x}^*_r, p)$, where:
\begin{equation} \label{eq:ss_hopf}
   \mathbf{x}^*_r \sim \mathcal{N}\left(\mathbf{x}_\mu^*,  \mathbf{J}_\mu^{-1}(\mathbf{x}_\mu^*,p) \mathbf{\Sigma}(\mathbf{x}_\mu^*,p)  \mathbf{J}_\mu^{-1}(\mathbf{x}_\mu^*,p)^\top  \right). 
\end{equation}

\noindent \textbf{Note}: $\frac{\partial \mathbf{B}}{\partial \mathbf{x}}$ is a tensor in $\mathbb{R}^{n\times n \times n}$.

\subsubsection{Jacobian distribution at steady states}
\label{subsubsec_app:eighopf}

\noindent In this subsection the goal is to estimate the distribution of Jacobian matrices corresponding to $n-$dimensional steady states $\mathbf{x}_r^*$, the distribution of which was estimated in Sec.~\ref{subsubsec_app:hopf_ss}. As a reminder, the solution branch in this case is expressed as $\mathbf{x}(p)$ and, for a fixed value of $p$, the steady states would be:  $(\mathbf{x}^*_r, p)$,  where $\mathbf{x}^*_r \sim \mathcal{N}\left(\mathbf{x}_\mu^*,  \mathbf{J}_\mu^{-1}(\mathbf{x}_\mu^*,p) \mathbf{\Sigma}(\mathbf{x}_\mu^*,p)  \mathbf{J}_\mu^{-1}(\mathbf{x}_\mu^*,p)^\top  \right)$ (see Eq.~\ref{eq:ss_hopf}).

\begin{align}
& \textcolor{black}{\text{Using  Eq.~\ref{eq:f_ndim} as a starting point}}: \notag \\[1ex]
%
%
%
& \textcolor{black}{\frac{\partial {f}_{r_i}}{\partial x_j}(\mathbf{x}_r^*, p) = \frac{\partial  \overline{f}_i}{\partial x_j}(\mathbf{x}_r^*,p) +  \frac{\partial B_{ik}}{\partial  x_j}(\mathbf{x}_r^*,p)\Theta_{r_k}}. \label{eq:jac_init} \\[1ex]
%
%
& \textcolor{black}{\text{Substituting $\mathbf{x}_r^* = \mathbf{x}_\mu^* + \mathbf{P}^*_r$:}}\notag \\[1ex]
%
%
%
& \textcolor{black}{\frac{\partial {f}_{r_i}}{\partial x_j}(\mathbf{x}_r^*, p) = \frac{\partial  \overline{f}_i}{\partial x_j}(\mathbf{x}_\mu^* + \mathbf{P}^*_r,p) +  \frac{\partial B_{ik}}{\partial  x_j}(\mathbf{x}_\mu^* + \mathbf{P}^*_r,p)\Theta_{r_k}}. \\[1ex]
& \textcolor{black}{\text{Expanding the functions $\frac{\partial {\overline{f}}_{i}}{\partial x_j}(\cdot, p), \frac{\partial B_{ik}}{\partial  x_j}(\cdot, p)$ around  $\mathbf{x}_\mu^*$ using Taylor series:}}  \notag \\[1ex]
%
%
%
&\textcolor{black}{\frac{\partial {f}_{r_i}}{\partial x_j}(\mathbf{x}_r^*, p)   = \frac{\partial  \overline{f}_i}{\partial x_j}(\mathbf{x}_\mu^*,p)+ \frac{\partial^2  \overline{f}_i}{\partial x_j\partial x_k}(\mathbf{x}_\mu^*,p)P_{r_k}^*+} \notag \\[1ex]
&\hspace{2.1cm} \textcolor{black}{\frac{\partial B_{ik}}{\partial  x_j}(\mathbf{x}_\mu^*,p)\Theta_{r_k}  +   \frac{\partial^2 B_{ik}}{\partial  x_j \partial x_l}(\mathbf{x}_\mu^*,p)P_{r_l}^*\Theta_{r_k}+ \mathcal{O}(||\mathbf{P}_{r}^*||^2_2)}. \\[1ex]
& \textcolor{black}{\text{Assuming again $||\mathbf{P}_{r}^*||_2$ is small, we can omit the remainder $\mathcal{O}(||\mathbf{P}_{r}^*||^2_2)$.}} \notag \\
& \textcolor{black}{\text{We can also eliminate the term containing elements of $\frac{\partial^2\mathbf{B}}{\partial  \mathbf{x}^2}(\mathbf{x}_\mu^*,p)\mathbf{P}^*_r\mathbf{\Theta}_{r}$}} \notag \\
& \text{where products of elements of $\mathbf{P}^*_{r},  \frac{\partial^2\mathbf{B}}{\partial  \mathbf{x}^2}(\mathbf{x}_\mu^*,p)$ appear:} \notag\\[1ex]
%
%
%
&\textcolor{black}{\frac{\partial {f}_{r_i}}{\partial x_j}(\mathbf{x}_r^*, p) \approx  \frac{\partial  \overline{f}_i}{\partial x_j}(\mathbf{x}_\mu^*,p)+ \frac{\partial^2  \overline{f}_i}{\partial x_j\partial x_k}(\mathbf{x}_\mu^*,p)P^*_{r_k}+  \frac{\partial B_{ik}}{\partial  x_j}(\mathbf{x}_\mu^*,p)\Theta_{r_k}}. \\[1ex]
%
%
%
%
& \textcolor{black}{\text{Then:}} \notag \\[1ex]
& E_r\left[\frac{\partial {f}_{r_i}}{\partial x_j}(\mathbf{x}_r^*, p) \right] = \frac{\partial {\overline{f}}_{i}}{\partial x_j}(\mathbf{x}^*_\mu,p) \hspace{0.2cm} \text{or} \hspace{0.2cm}E_r[\mathbf{J}_r(\mathbf{x}^*_r,p) ] = \mathbf{J}_\mu(\mathbf{x}^*_\mu,p) \label{eq:meanjac}, \\[1ex]
%
%
%
&\mathbf{\Sigma_{J_r}} \equiv \mathrm{Cov}(\mathbf{J}_r(\mathbf{x}^*_r,p))  = \text{(See property~\ref{prop:3} for elementwise calculations)} \label{eq:varjac} 
%
%
%
\end{align}


\subsubsection{(Squared) eigenvalue distribution}
\label{subsubsec_app:eigv}

Having estimated the distribution of Jacobian matrices at the steady states (Eqs.~\ref{eq:meanjac}, \ref{eq:varjac}), now we examine how this translates to the distribution of their respective eigenvalues. For simplicity, we will deal with a single eigenvalue (out of a total number $n$), which we assume suffices for the detection of a Hopf bifurcation. Alternatively, one could track the statistics of the trace of the Jacobian, which can be easily obtained from the distribution of the Jacobian (see property~\ref{prop:4}).\\
\noindent Notation relevant to this subsection:

\begin{itemize}
    \renewcommand{\labelitemi}{\scriptsize$\blacksquare$}
    \item $\lambda_\mu \in \mathbb{C}$ is the eigenvalue of interest of $\mathbf{J}_\mu (\mathbf{x}_\mu^*,p)$ (at the steady state of the predictive mean).
    \item $\lambda_r \in \mathbb{C}$ is the eigenvalue of interest of $\mathbf{J}_r(\mathbf{x}_r^*,p)$ (at the steady state of the GP realization indexed $r$).
    \item $\mathbf{v_R, v_L} \in \mathbb{C}^n$ are the right and left eigenvectors of $\mathbf{J}_{\mu}(\mathbf{x}_\mu^*,p)$ respectively, corresponding to the eigenvalue of interest.
\end{itemize}

Regarding the eigenvalue statistics, assuming that $\mathbf{J}_r(\mathbf{x}_r^*,p) = \mathbf{J}_\mu(\mathbf{x}_\mu^*,p)+\rho_r^J$:

\begin{align}
   & \lambda(\mathbf{J}_r(\mathbf{x}^*_r,p)) = \lambda(\mathbf{J}_\mu(\mathbf{x}^*_\mu,p)+\rho_r^J).\\[1ex]
   & \textcolor{black}{\text{Writing the above elementwise and dropping the arguments for simplicity:}} \notag \\[1ex]
   & \lambda(J_{r_{ij}}) = \lambda(J_{\mu_{ij}}+\rho_{r_{ij}}^J).\\[1ex]
   & \textcolor{black}{\text{Using Taylor expansion for } \lambda(\cdot): } \notag \\[1ex]
   & \lambda(J_{r_{ij}}) = \lambda(J_{\mu_{ij}})+\frac{\partial \lambda}{\partial J_{ij}}(J_{\mu_{ij}})\rho_{r_{ij}}^J+ \mathcal{O}({\rho_{r_{ij}}^J}^2).\\[1ex]
   & \textcolor{black}{\text{Using the results of Lancaster \cite{LANCASTER1964} (specifically, Eq.~5): }} \notag \\[1ex]
   & \lambda(J_{r_{ij}}) = \lambda(J_{\mu_{ij}})+ \frac{1}{\mathbf{\overline{v_L}}\mathbf{v_R}} \mathbf{\overline{v_L}} \frac{\partial\mathbf{J}}{\partial J_{ij}}(J_{\mu_{ij}})\mathbf{v_R} \rho_{r_{ij}}^J+ \mathcal{O}({\rho_{r_{ij}}^J}^2).\\[1ex]
   & \lambda(J_{r_{ij}}) = \lambda(J_{\mu_{ij}})+ \frac{1}{\overline{v_L}_i{v_R}_j} \overline{v_L}_i{v_R}_j \rho_{r_{ij}}^J+ \mathcal{O}({\rho_{r_{ij}}^J}^2),\\[1ex]
   & \textcolor{black}{\text{where overline denotes complex conjugate.}} \notag\\
   &\text{Summing the contributions of all elements of the Jacobian, results in:}\notag \\
   %
   %
   %
   %
   & \lambda(\mathbf{J}_{r}) = \lambda(\mathbf{J}_{\mu})+ \frac{1}{\mathbf{\overline{v_L}}\mathbf{v_R}} \mathbf{\overline{v_L}}\rho_r^J\mathbf{v_R}+ \mathcal{O}(||\rho_{r}^J||^2). \label{eq:eig1} \\[1ex]
   & \textcolor{black}{\text{In Eq.~\ref{eq:eig1}, $\rho_r^J \in \mathbb{R}^{n \times n}$. Consider $\rho_{vec_r}^J \in \mathbb{R}^{n^2}$ the vectorized version of $\rho_r^J$.}} \notag \\
   &\textcolor{black}{\text{Also consider $\mathbf{v}_{aux} = [...,v_{L_{1+j mod n}}v_{R_{1+\lfloor j \rfloor }},...] \in \mathbb{R}^{n^2}$. Then Eq.~\ref{eq:eig1} can be written: }} \notag \\[1ex]
   & \lambda(\mathbf{x}_r^*,p) = \lambda_\mu+ \frac{1}{\mathbf{\overline{v_L}}\mathbf{v_R}} \mathbf{v}_{aux}^\top\rho_{vec_r}^J +\mathcal{O}(||\rho_{vec_r}^J||_2^2). \label{eq:eig2} \\[1ex]
   & \textcolor{black}{\text{Therefore :}} \notag \\[1ex]
   &E_r[\lambda(\mathbf{J}_{r}) ] = \lambda_\mu \equiv \lambda(\mathbf{J}_{\mu}),  \\[1ex]
   &Var(\lambda(\mathbf{J}_{r}) ) = Var \left( \frac{1}{\mathbf{\overline{v_L}}\mathbf{v_R}} \mathbf{v}_{aux}^\top\rho_{vec_r}^J\right) = \frac{1}{(\mathbf{\overline{v_L}}\mathbf{v_R})^2} \mathbf{v}_{aux}^\top \mathbf{\Sigma_{J_r}} \mathbf{v}_{aux} + \mathcal{O}(||\rho_{vec_r}^J||_2^4).\\[1ex]
   & \textcolor{black}{\text{Or in Einstein notation: }} \notag \\[1ex]
   &\textcolor{black}{Var(\lambda_r) = \frac{1}{(v_{L_i} v_{R_i})^2} v_{L,i_1} v_{L,i_2} v_{R,j_1} v_{R,j_2} \Sigma_{J_{r_{i_1i_2j_1j_2}}} + \mathcal{O}(||\rho_{vec_r}^J||^4) },
\end{align}

\noindent as $\rho_{vec_r}^J(\rho_{vec_r}^J)^\top = \mathbf{\Sigma_{J_r}}$ and $i_1, i_2, j_1, j_2 \in \{0,...,n\}$.\\

To discover Hopf bifurcations, only the real part of an eigenvalue is of interest (assuming we are looking at the eigenvalue with the minimum absolute real part). The statistics of the real part of the eigenvalue, starting from Eq.~\ref{eq:eig2} would be:

\begin{align}
    & Re( \lambda_r(\mathbf{x}_r^*,p)) = Re(\lambda_\mu)+ Re\left(\frac{1}{\mathbf{\overline{v_L}}\mathbf{v_R}} \mathbf{v}_{aux}^\top\rho_{vec_r}^J\right),  \\[1ex]
    & Re( \lambda_r(\mathbf{x}_r^*,p)) = Re(\lambda_\mu)+ Re\left(\frac{1}{\mathbf{\overline{v_L}}\mathbf{v_R}} \mathbf{v}_{aux}^\top\right)\rho_{vec_r}^J. \\[1ex]
    & Re( \lambda_r(\mathbf{x}_r^*,p)) = Re(\lambda_\mu)+ \frac{1}{|\mathbf{\overline{v_L}}\mathbf{v_R}|^2}(Re( \mathbf{v}_{aux}^\top)Re(\mathbf{\overline{v_L}}\mathbf{v_R})  +Im( \mathbf{v}_{aux}^\top)Im(\mathbf{\overline{v_L}}\mathbf{v_R}))\rho_{vec_r}^J. \\[1ex]
    & \textcolor{black}{\text{Setting $\mathbf{z} = Re( \mathbf{v}_{aux}^\top)Re(\mathbf{\overline{v_L}}\mathbf{v_R})  +Im( \mathbf{v}_{aux}^\top)Im(\mathbf{\overline{v_L}}\mathbf{v_R})$:}} \label{eq:def_z} \\[1ex]
    & E[Re( \lambda_r(\mathbf{x}_r^*,p))] =  Re(\lambda_\mu),   \\[1ex]
    & Var(Re( \lambda_r(\mathbf{x}_r^*,p))) =\frac{1}{|\mathbf{\overline{v_L}}\mathbf{v_R}|^4} \mathbf{z}\mathbf{\Sigma_{J_r}}{\mathbf{z}^\top}.
\end{align}

For the distribution of the \textbf{squared} real part of an eigenvalue of interest we can again use the noncentral Gaussian moments (see Property 2) to get:

\begin{align}
&E\left[Re( \lambda_r(\mathbf{x}_r^*,p))^2\right] = E\left[Re( \lambda_r(\mathbf{x}_r^*,p))\right]^2 + Var\left(Re( \lambda_r(\mathbf{x}_r^*,p))\right),\\[1ex]
& Var\left(Re( \lambda_r(\mathbf{x}_r^*,p))^2\right) = 2Var\left(Re( \lambda_r(\mathbf{x}_r^*,p))\right) \left(2 E\left[Re( \lambda_r(\mathbf{x}_r^*,p))\right]^2 + Var\left(Re( \lambda_r(\mathbf{x}_r^*,p))\right)  \right).
\end{align}

\subsection{Properties}


\begin{enumerate}


\item $E[X^2] = Var(X)+E[X]^2$,

\hspace{-0.2cm} $Var(X^2) = 2Var(X)(Var(X)+2E[X]^2)$   \label{prop:1}\\

Proof:\\
For $E[X^2]$, this is straight-forward from the definition of variance.
For the variance, assuming $X \sim \mathcal{N}(\mu, \sigma^2) $ we can use the moment generating function of the normal distribution to show that $E[X^4] = \mu^4 + 6\mu^2\sigma^2 + 3\sigma^4$. Then:\\

$Var(X^2) = E[X^4] - E[X^2]^2 = \mu^4 + 6\mu^2\sigma^2 + 3\sigma^4 - (\sigma^2+ \mu^2)^2 = 2\sigma^2(\sigma^2+2\mu^2).$\\





\item Element-wise Jacobian covariance:  \label{prop:3}
\begin{align}
    &\mathrm{Cov} \left(\frac{\partial f_{r_{i_1}}}{\partial x_{j_1}}(\mathbf{x}_r^*,p), \frac{\partial f_{r_{i_2}}}{\partial x_{j_2}}(\mathbf{x}_r^*,p)\right)= \\
    &\mathrm{Cov} \biggl( \frac{\partial  \overline{f}_{i_1}}{\partial x_{j_1}}(\mathbf{x}_\mu^*,p)+ \frac{\partial^2  \overline{f}_{i_1}}{\partial x_{j_1}\partial x_k}(\mathbf{x}_\mu^*,p)P^*_{r_k}+  \frac{\partial B_{i_1k}}{\partial  x_{j_1}}(\mathbf{x}_\mu^*,p)\Theta_{r_k},\notag\\
    &\hspace{0.9cm} \frac{\partial  \overline{f}_{i_2}}{\partial x_{j_2}}(\mathbf{x}_\mu^*,p)+ \frac{\partial^2  \overline{f}_{i_2}}{\partial x_{j_2}\partial x_k}(\mathbf{x}_\mu^*,p)P^*_{r_k}+  \frac{\partial B_{i_2k}}{\partial  x_{j_2}}(\mathbf{x}_\mu^*,p)\Theta_{r_k}\biggl )= \\
    &\mathrm{Cov} \biggl(\frac{\partial^2  \overline{f}_{i_1}}{\partial x_{j_1}\partial x_k}(\mathbf{x}_\mu^*,p)P^*_{r_k}+  \frac{\partial B_{i_1k}}{\partial  x_{j_1}}(\mathbf{x}_\mu^*,p)\Theta_{r_k},\notag\\
    &\hspace{0.9cm} \frac{\partial^2  \overline{f}_{i_2}}{\partial x_{j_2}\partial x_k}(\mathbf{x}_\mu^*,p)P^*_{r_k}+  \frac{\partial B_{i_2k}}{\partial  x_{j_2}}(\mathbf{x}_\mu^*,p)\Theta_{r_k}\biggl )=\\
    & \frac{\partial^2  \overline{f}_{i_1}}{\partial x_{j_1}\partial x_{k_1}}(\mathbf{x}_\mu^*,p) \frac{\partial^2  \overline{f}_{i_2}}{\partial x_{j_2}\partial x_{k_2}}(\mathbf{x}_\mu^*,p) \Sigma^*_{k_1k_2} + 
    \frac{\partial B_{i_1k}}{\partial  x_{j_1}}(\mathbf{x}_\mu^*,p)\frac{\partial B_{i_2k}}{\partial  x_{j_2}}(\mathbf{x}_\mu^*,p) \notag \\
    & -\frac{\partial^2  \overline{f}_{i_1}}{\partial x_{j_1}\partial x_{k_1}}(\mathbf{x}_\mu^*,p) J^{-1}_{\mu_{k_1m}}(\mathbf{x}_\mu^*,p)B_{ml}(\mathbf{x}_\mu^*,p)\frac{\partial B_{i_2l}}{\partial  x_{j_2}}(\mathbf{x}_\mu^*,p) \notag \\
    &-\frac{\partial^2  \overline{f}_{i_2}}{\partial x_{j_2}\partial x_{k_2}}(\mathbf{x}_\mu^*,p) J^{-1}_{\mu_{k_2m}}(\mathbf{x}_\mu^*,p)B_{ml}(\mathbf{x}_\mu^*,p)\frac{\partial B_{i_1l}}{\partial  x_{j_1}}(\mathbf{x}_\mu^*,p).
    %
    %
\end{align}

The above formula was derived using the following properties/assumptions:

\begin{itemize}
    \item $\mathrm{Cov} \bigl( \sum_{i} a_i X_i , \sum_{j} b_j Y_j \bigl) = \sum_{i,j} a_ib_j\mathrm{Cov}(X_i, Y_j)$.
    \item $\mathrm{Cov}(\Theta_{r_{l_1}}, \Theta_{r_{l_2}}) = \delta_{l_1l_2}$.
    \item $\mathrm{Cov}(P^*_{r_{l_1}}, P^*_{r_{l_2}}) = \Sigma^*_{l_1l_2}$.
    \item $\mathrm{Cov}(P^*_{r_{l_1}}, \Theta_{r_{l_2}}) = - J^{-1}_{\mu_{l_1m}}B_{ml_2}$.
    %
\end{itemize}
where $\delta_{ij}$ is the Kronecker delta. 

\item Distribution of the trace of the Jacobian.  \label{prop:4}
\begin{align}
    & E[tr(\mathbf{J}_r(\mathbf{x}^*_r,p)] = E[\Sigma_{i=1}^n J_{r_{ii}}(\mathbf{x}^*_r,p)] = 
    \Sigma_{i=1}^n J_{\mu_{ii}}(\mathbf{x}^*_\mu,p), \\[1ex]
    & Var(tr(\mathbf{J}_r(\mathbf{x}^*_r,p)) = Var(\Sigma_{i=1}^n J_{r_{ii}}(\mathbf{x}^*_r,p)) =  \sum_{i=1}^n\sum_{j=1}^n\mathrm{Cov}(J_{r_{ii}}(\mathbf{x}^*_r,p), J_{r_{jj}}(\mathbf{x}^*_r,p)). \label{eq:var_trace}
\end{align}

Note that all components of Eq.~\ref{eq:var_trace} are elements of $\mathbf{\Sigma_{J_r}}$ which are already calculated.


\end{enumerate}

\section{Algorithms}
\label{sec_app:algos}

\subsection{Uncertainty Sampling}
\label{subsec_app:us}

The following algorithm can be used for 1D uncertainty sampling. It can also be adapted for higher-dimensional vectorfields e.g. by considering sums of variances.

\begin{algorithm}[H]

\caption{Uncertainty Sampling in 1D}\label{alg:us}

\textbf{Input}:\\ 
Surrogate model: $\hat{g}\approx f (x;p)$ \\
Grid of points in the input space (state and parameter): $(x,p)_{i=1}^{N_{grid}}$

\textbf{Output}: \\ New candidate point  ($x^{new}, p^{new})$
\vspace{0.2cm}

\begin{algorithmic}

\State  Calculate the variance at every point on the input space: $Var(\hat{g})_i \gets Var(\hat{g}(x_i, p_i)) \hspace{0.2cm} i =1, ..., N_{grid}$

\State Choose point with maximum uncertainty: $(x^{new}, p^{new}) \gets (x_I, p_I),\hspace{0.2cm} I= argmax_i Var(\hat{g})_i$

\end{algorithmic}
\end{algorithm}

\subsection{Monte Carlo acquisition function}
\label{subsec_app:mc}

\begin{algorithm}[H]

\caption{Monte Carlo acquisition function for Hopf bifurcations in 2D}\label{alg:nd_hopf_mc}

\textbf{Input}:\\ 
Surrogate model: $\mathbf{\hat{g}}\approx \mathbf{f} (\mathbf{x};p)$\\
Monte Carlo sample: $N_{MC}$\\

\textbf{Output}: \\ Monte Carlo acquisition function value: $\alpha_{MC}(p; \mathbf{\hat{g}})$
\vspace{0.2cm}

\begin{algorithmic}

\Procedure{Acquisition function: $\alpha_{MC}(p; \mathbf{\hat{g}})$}{}
\State Sample $N_{MC}$ realizations of the probabilistic surrogate, indexed by $r$: $\mathbf{\hat{g}}_r, \hspace{0.2cm} r=1,..., N_{MC}$

\State Given an initial state $\mathbf{x}_0$, numerically solve each $\mathbf{\hat{g}}_r$ for the steady state.  
\State  \hspace{1cm} $\mathbf{x}^*_r(p) \xleftarrow{Newton} \mathbf{x}_0,p,\mathbf{\hat{g}}_r, \hspace{0.5cm} r=1,...,N_{MC}$

\State For each realization of the surrogate, indexed $r$, calculate the Jacobian at the steady state.

\State \hspace{1cm} $\mathbf{J}_r(\mathbf{x}^*_r;p) \gets \mathbf{x}^*_r,p, \mathbf{\hat{g}}_r, \hspace{0.5cm} r=1,...,N_{MC}$

\State  For each realization of the surrogate, indexed $r$, calculate the eigenvalue set at the steady state,
\State given the Jacobian.

\State \hspace{1cm} $(\lambda_{r_1}, ...\lambda_{r_n})  \gets \mathbf{J}_r(\mathbf{x}^*_r;p), \hspace{0.5cm} r=1,...,N_{MC}$

\State  For each eigenvalue set, choose the one with the least absolute real part ($\lambda_{r}$). 
\State Then, calculate the statistics of the resulting eigenvalue sample.

\State \hspace{1cm} $\overline{\lambda_{r}}, std(\lambda_{r}) \gets \{ \lambda_r \}_{r=1}^{N_{MC}}$

\State Then, calculate the statistics of the resulting sample of squared eigenvalues.

\State \hspace{1cm} $\overline{\lambda_{r}^2}, std(\lambda_{r}^2) \gets \overline{\lambda_{r}}, std(\lambda_{r})$ 

\State  Calculate a standard acquisition function $\alpha_{MC}(p; \mathbf{\hat{g}})$, using the above statistics,
\State e.g. Lower Confidence bound:
\State \hspace{1cm} $\alpha_{MC}(p; \mathbf{\hat{g}}) = \overline{\lambda_{r}^2}-\beta std(\lambda_{r}^2)$
\EndProcedure

\end{algorithmic}
\end{algorithm}







\bibliographystyle{siam}
\bibliography{bib, library}

\end{document}